\documentclass[10pt,twocolumn,letterpaper]{article}

\usepackage{iccv}
\usepackage{times}
\usepackage{epsfig}
\usepackage{graphicx}
\usepackage{amsmath}
\usepackage{amssymb}

\usepackage{amssymb}
\usepackage{bbm} 
\usepackage{mathtools} 
\usepackage{enumitem}
\usepackage{array,multirow}
\usepackage{subcaption}
\usepackage{pifont}
\usepackage{booktabs}
\usepackage{diagbox}
\usepackage{hhline}
\usepackage[export]{adjustbox}
\usepackage[numbers,sort]{natbib}
\usepackage{pdfpages}
\usepackage[accsupp]{axessibility}  

\usepackage[pagebackref=true,breaklinks=true,letterpaper=true,colorlinks,bookmarks=false]{hyperref}

\newcommand{\fm}{\footnotemark}
\newcommand{\rb}{\rotatebox[origin=c]{90}}%
\newcolumntype{L}[1]{>{\raggedright\let\newline\\\arraybackslash\hspace{0pt}}m{#1}}
\newcolumntype{R}[1]{>{\raggedleft\let\newline\\\arraybackslash\hspace{0pt}}m{#1}}
\newcolumntype{C}[1]{>{\centering\let\newline\\\arraybackslash\hspace{0pt}}m{#1}}

\iccvfinalcopy 

\def\iccvPaperID{10410} 

\ificcvfinal\fi

\begin{document}

\title{ACLS: Adaptive and Conditional Label Smoothing for Network Calibration}

\author{%
   Hyekang Park$^{1}$ \hspace{8mm} Jongyoun Noh$^{1}$ \hspace{8mm} Youngmin Oh$^{1}$                        \\
   Donghyeon Baek$^{1}$ \hspace{8mm} {Bumsub Ham$^{1,2}$\thanks{Corresponding author}} \vspace{2.5mm}         \\ 
   $^{1}$Yonsei University \hspace{8mm} $^{2}$Korea Institute of Science and Technology~(KIST) \vspace{2mm} \\ 
   {\url{https://cvlab.yonsei.ac.kr/projects/ACLS}}
}

\maketitle
\ificcvfinal\thispagestyle{empty}\fi

\begin{abstract}
We address the problem of network calibration adjusting miscalibrated confidences of deep neural networks. Many approaches to network calibration adopt a regularization-based method that exploits a regularization term to smooth the miscalibrated confidences. Although these approaches have shown the effectiveness on calibrating the networks, there is still a lack of understanding on the underlying principles of regularization in terms of network calibration. We present in this paper an in-depth analysis of existing regularization-based methods, providing a better understanding on how they affect to network calibration. Specifically, we have observed that 1) the regularization-based methods can be interpreted as variants of label smoothing, and 2) they do not always behave desirably. Based on the analysis, we introduce a novel loss function, dubbed ACLS, that unifies the merits of existing regularization methods, while avoiding the limitations. We show extensive experimental results for image classification and semantic segmentation on standard benchmarks, including CIFAR10, Tiny-ImageNet, ImageNet, and PASCAL VOC, demonstrating the effectiveness of our loss function. 
\end{abstract}


\vspace{-2mm}
\section{Introduction}
\label{sec:intro}

Humans have an ability to make well-calibrated decisions, such that confidence levels of decisions reflect likelihoods of underlying events accurately. On the contrary, deep neural networks (DNNs) often struggle to achieve such levels of calibration, which is problematic particularly in applications involving high levels of uncertainty and reasoning, including autonomous driving~\cite{grigorescu2020survey,stocco2020misbehaviour,helldin2013presenting} and medical diagnosis~\cite{jiang2012calibrating,mehrtash2020confidence,carneiro2020deep}. For example, supervised approaches to image classification typically exploit one-hot encoded labels and a softmax cross-entropy loss~\cite{guo2017calibration,mukhoti2020calibrating,cheng2022calibrating,liu2022devil} to train the networks. The loss encourages minimizing the entropy of network outputs,~\ie, preferring the Dirac delta distribution with a peak at the one-hot label, which causes overconfident predictions, while overly penalizing uncertainty in the predictions~\cite{liu2022devil}.

\begin{table}[]
   \captionsetup{font={small}}
   \caption{Comparison of regularization-based methods for network calibration. Expected calibration error (ECE) is computed with 15 bins using ResNet-50~\cite{he2016deep} on Tiny-ImageNet~\cite{deng2009imagenet}. We denote by $\bigtriangleup$ adaptive or conditional regularizers with negative effects. LS: Label smoothing.}
   \vspace{-1mm}
   \label{tab:teaser}
   \centering
   \renewcommand{\arraystretch}{1.0}
   \setlength{\tabcolsep}{2.2pt}
   \resizebox{\linewidth}{!}{%
   \begin{tabular}{l|c|c|c} \hline
   \multicolumn{1}{c|}{\multirow{2}{*}{Method}} &
     \multirow{2}{*}{\begin{tabular}[c]{@{}c@{}}Adaptive    \\ Regularization (AR)\end{tabular}} &
     \multirow{2}{*}{\begin{tabular}[c]{@{}c@{}}Conditional \\ Regularization (CR)\end{tabular}} &
     \multirow{2}{*}{ECE$\downarrow$ (\%)}                  \\ 
                                                               &                    &                  &                \\ \hline\hline
   LS~\cite{szegedy2016rethinking}                             &     -              &    -             & 3.17           \\
   FLSD~\cite{mukhoti2020calibrating}                          &     -              &    -             & 2.91           \\
   CPC~\cite{cheng2022calibrating}                             & $\bigtriangleup$   &    -             & 3.12           \\
   MDCA~\cite{hebbalaguppe2022stitch}                          & $\bigtriangleup$   &    -             & 2.77           \\
   MbLS~\cite{liu2022devil}                                    &     -              & $\bigtriangleup$ & 1.64           \\
   CRL~\cite{moon2020confidence}                               & $\bigtriangleup$   & $\bigtriangleup$ & 1.65           \\ \hline
   ACLS                                                        & $\checkmark$       & $\checkmark$     & \textbf{1.05}  \\ \hline
   \end{tabular}
   }
   \vspace{-5mm}
\end{table}

Recently, a variety of confidence calibration methods have been introduced to address this problem~\cite{mukhoti2020calibrating,cheng2022calibrating,liu2022devil,hebbalaguppe2022stitch,moon2020confidence,guo2017calibration,ghosh2022adafocal,muller2019does,ding2021local,ma2021meta}, which can broadly be divided into two categories: Post-hoc methods and regularization-based methods. Post-hoc approaches to network calibration adjust the predictions of a pre-trained model at test time, typically using additional trainable parameters~\cite{platt1999probabilistic,guo2017calibration,ding2021local}. For example, a temperature scaling technique~\cite{platt1999probabilistic} multiplies logits by a temperature parameter, resulting in a softened distribution of predictions that can help mitigate the overconfidence problem. While post-hoc approaches are effective and computationally cheap, they mainly have two limitations. First, they require a held-out dataset to tune the additional parameters~\cite{hebbalaguppe2022stitch,mukhoti2020calibrating}. Second, post-hoc approaches assume that training and test samples are drawn from the same distribution~\cite{liu2022devil}, which limits performance under distribution shifts between training and test datasets~\cite{kumar2018trainable,mukhoti2020calibrating,hebbalaguppe2022stitch}. Regularization approaches integrate calibration techniques into the training process,~\eg,~in a form of objective functions~\cite{liu2022devil,hebbalaguppe2022stitch,moon2020confidence,cheng2022calibrating,mukhoti2020calibrating}. These methods regularize network outputs implicitly~\cite{mukhoti2020calibrating,muller2019does} or explicitly~\cite{liu2022devil,hebbalaguppe2022stitch,moon2020confidence,cheng2022calibrating} by penalizing overconfident or underconfident predictions, encouraging the distribution of predictions to be uniform. While regularization approaches have shown the effectiveness, the influence of regularization terms on network calibration remains unclear, and only a limited number of studies have explored to delve deeper into these approaches. For instance, the works of~\cite{muller2019does,liu2022devil} have demonstrated that the label smoothing technique~\cite{szegedy2016rethinking} is also effective for network calibration. In addition, the focal loss~\cite{lin2017focal}, which was initially developed for object detection, has been proven to raise the entropy of network predictions, performing regularization implicitly~\cite{mukhoti2020calibrating,liu2022devil}.

We present in this paper a theoretical and empirical analysis of various regularization-based methods, including LS~\cite{szegedy2016rethinking}, FLSD~\cite{mukhoti2020calibrating}, CPC~\cite{cheng2022calibrating}, MDCA~\cite{hebbalaguppe2022stitch}, MbLS~\cite{liu2022devil}, and CRL~\cite{moon2020confidence}, to better understand the underlying principles of these approaches. Our analysis on the gradients of objective functions for the regularization-based methods reveals that 1) these methods can be viewed as variants of the label smoothing technique~\cite{szegedy2016rethinking}, and differ only in how they determine the degree of label smoothing, and 2) they do not always behave as expected, and often produce undesired results in terms of network calibration. We further categorize the regularization-based methods into three groups based on the type of label smoothing: Adaptive regularization~(AR)~\cite{cheng2022calibrating,hebbalaguppe2022stitch}, conditional regularization~(CR)~\cite{liu2022devil}, and a combination of them~\cite{moon2020confidence}. AR adjusts the strength of regularization for (one-hot encoded) training labels adaptively based on output probabilities of a network across classes, which is more beneficial for network calibration than the vanilla label smoothing technique~(CPC~\cite{cheng2022calibrating}, MDCA~\cite{hebbalaguppe2022stitch} vs. LS~\cite{szegedy2016rethinking} in Table~\ref{tab:teaser}). Ideally, the label of a true class should decrease in accordance with the degree of increase in a corresponding output probability, whereas the labels of false classes should increase proportionally with the degree of decrement in each output probability. However, our observation suggests that the regularization-based methods using AR~\cite{cheng2022calibrating,hebbalaguppe2022stitch} do not consistently behave as in the ideal scenario. Specifically, when the output probability is exceedingly high, the label of a true class rather decreases slightly. CR modifies training labels selectively based on a specific criterion,~\eg,~using margin-based penalties~\cite{liu2022devil}. Since output probabilities of a network are not always miscalibrated, CR performs regularization on the miscalibrated ones only, thus showing better calibration capability than the label smoothing technique~(MbLS~\cite{liu2022devil} vs. LS~\cite{szegedy2016rethinking} in Table~\ref{tab:teaser}). However, MbLS~\cite{liu2022devil} using CR is not likely to regularize the training label of a true class, which leads to the overconfidence problem. A hybrid method,~\eg,~CRL~\cite{moon2020confidence}, takes advantages of AR and CR, but it would inherit the limitations of both approaches. We will provide a more detailed analysis in Sec.~\ref{sec:3.2}.

Based on the gradient analysis for the regularization-based approaches, we introduce a novel loss function, dubbed ACLS, for network calibration. ACLS combines the strengths of AR and CR, while mitigating the drawbacks of the regularization-based methods~\cite{cheng2022calibrating,hebbalaguppe2022stitch,liu2022devil,moon2020confidence,lin2017focal,mukhoti2020calibrating,szegedy2016rethinking}, providing better calibration results~(Table~\ref{tab:teaser}). On the one hand, it determines the degree of label smoothing adaptively for each class, while avoiding the undesirable aspects observed in the regularization-based methods using AR~\cite{cheng2022calibrating,hebbalaguppe2022stitch}. Specifically, ACLS regularizes the labels of true object classes more strongly, as corresponding output probabilities increase, and vice versa for other classes. On the other hand, it exploits a predefined margin to determine whether to adjust training labels, similar to the regularization method using CR~\cite{liu2022devil}, but also modifies the label of a true class to smooth a corresponding probability, preventing the overconfidence problem. Experimental results on standard benchmarks~\cite{deng2009imagenet,krizhevsky2009learning,le2015tiny,everingham2009pascal} demonstrate that networks trained with ACLS outperform the state of the art in terms of expected calibration error~(ECE) and adaptive ECE~(AECE). Our main contributions can be summarized as follows:

\vspace{-1.5mm}
\begin{itemize}[leftmargin=*]
   \item[$\bullet$] We present an in-depth analysis of existing loss functions for network calibration~\cite{cheng2022calibrating,hebbalaguppe2022stitch,liu2022devil,moon2020confidence,lin2017focal,mukhoti2020calibrating,szegedy2016rethinking}. We show that current calibration methods can be viewed as variations of the label smoothing technique~\cite{szegedy2016rethinking}, and they are limited to prevent overconfidence and/or underconfidence problems. 
   \vspace{-2mm}
   \item[$\bullet$] Based on the analysis, we present a new loss function, ACLS, that retains the advantages of AR and CR, while overcoming the negative effects of existing calibration methods.   
   \vspace{-2mm}
   \item[$\bullet$] We set a new state of the art on standard benchmarks~\cite{deng2009imagenet,krizhevsky2009learning,le2015tiny}, including CIFAR10~\cite{krizhevsky2009learning}, Tiny-ImageNet~\cite{le2015tiny}, ImageNet~\cite{deng2009imagenet}, and PASCAL VOC~\cite{everingham2009pascal}, demonstrating the effectiveness of our method with extensive experiments and ablation studies.
   \vspace{-1mm}
\end{itemize}

\vspace{-2mm}
\section{Related work}

\vspace{-1mm}
\paragraph{Post-hoc calibration.}
A pioneer work of~\cite{guo2017calibration} proposes to calibrate predictions of a pre-trained DNN for image classification at test time, using a temperature scaling technique~\cite{platt1999probabilistic}. It regularizes the distribution of the predictions in a way of minimizing negative log-likelihood or ECE using a temperature parameter. ETS~\cite{zhang2020mix} aggregates various temperature scaling techniques to improve the calibration performance. Instead of using ensemble, the work of~\cite{tomani2022parameterized} proposes to estimate temperatures for individual samples adaptively. A recent work of~\cite{ding2021local} extends the temperature scaling technique for dense prediction tasks~(\eg,~semantic segmentation). Although these calibration methods are straightforward and effective, they reduce all output probabilities of a network, without considering confidence levels of individual predictions. It is thus highly likely that well-calibrated predictions could also be altered, which rather degrades the overall calibration performance~\cite{kumar2018trainable,mukhoti2020calibrating,hebbalaguppe2022stitch}. Moreover, tuning hyperparameters~(\eg,~the temperature) requires additional held-out datasets, which makes it challenging to deploy post-hoc calibration methods in practical settings~\cite{hebbalaguppe2022stitch}.

\vspace{-3mm}
\paragraph{Regularization-based calibration.}
Many attempts to calibrating neural networks have been made using regularization-based approaches. The seminal work of~\cite{pereyra2017regularizing} suppresses overconfident predictions, while maximizing the entropy of network outputs. FLSD~\cite{mukhoti2020calibrating} proposes to exploit the focal loss~(FL)~\cite{lin2017focal} for network calibration. In particular, it adjusts a focusing parameter adaptively to concentrate more on hard examples, alleviating the overconfidence problem. However, the FL ignores easy training samples, which causes the underconfidence problem~\cite{ghosh2022adafocal}. FLSD also uses heuristics to determine the focusing parameter which is not optimal for all samples. AdaFocal~\cite{ghosh2022adafocal} addresses these problems by setting the focusing parameter for each sample based on a calibration metric and exploiting an inverse focal loss~\cite{wang2021rethinking} to put more focus on easy samples. MbLS~\cite{liu2022devil} proposes to exploit margin-based penalties to selectively regularize miscalibrated predictions. Recently, the works of~\cite{cheng2022calibrating,hebbalaguppe2022stitch,moon2020confidence} present new regularization terms for network calibration. Specifically, CRL~\cite{moon2020confidence} formulates network calibration as an ordinal ranking problem~\cite{corbiere2019addressing}, and determines whether to penalize predictions of a network in order to obtain better confidence estimates. CPC~\cite{cheng2022calibrating} decouples multi-class predictions into multiple binary ones, and calibrates pairs of binary predictions, augmenting the number of supervisory signals, compared to the softmax cross-entropy loss. MDCA~\cite{hebbalaguppe2022stitch} introduces a differentiable version of static calibration error~(SCE)~\cite{nixon2019measuring}, enabling optimizing a network directly in terms of SCE.
   
Although regularization-based methods have proven effective in calibrating networks, there have been limited efforts to explore the underlying principles behind these approaches. The work of~\cite{mukhoti2020calibrating} offers both theoretical and empirical justifications of using the focal loss~\cite{lin2017focal} for calibrating networks. In particular, it shows that the focal loss minimizes implicitly the Kullback-Leibler~(KL) divergence between a uniform distribution and softmax predictions, regularizing the distribution of network predictions. The label smoothing technique~\cite{szegedy2016rethinking}, which improves the discriminative ability of DNNs, has been demonstrated to be effective for network calibration~\cite{muller2019does,liu2022devil}. Similar to the finding in~\cite{mukhoti2020calibrating}, the work of~\cite{liu2022devil} highlights that a cross-entropy loss with the label smoothing augments the KL divergence between a uniform distribution and softmax predictions. It also demonstrates that the focal loss can be regarded as a form of label smoothing, which effectively alleviates the overconfidence problem. Taking one step further, we show that existing regularization-based methods, including FLSD~\cite{mukhoti2020calibrating}, CPC~\cite{cheng2022calibrating}, MDCA~\cite{hebbalaguppe2022stitch}, MbLS~\cite{liu2022devil}, and CRL~\cite{moon2020confidence}, can be considered as variants of the label smoothing technique. We also provide a comprehensive analysis on gradients of objective functions for these methods, and show that the objective functions have detrimental effects on network calibration.

\section{Method} \label{sec:3}
\begin{table*}[]
   \captionsetup{font={small}}
   \caption{Gradient analysis of existing regularization-based methods for network calibration~\cite{cheng2022calibrating,hebbalaguppe2022stitch,liu2022devil,moon2020confidence,lin2017focal,mukhoti2020calibrating}. We compute the gradients of loss functions for the calibration methods w.r.t a logit, and reformulate the results as in Eq.~\eqref{eq:grad_unified}, where each method differs in how smoothing and indicator functions,~$f$ and $\mathbbm{C}$, are defined. Note that $\lambda_1$ and $\lambda_2$ are hyperparameters for each method. In CRL~\cite{moon2020confidence}, $H(n, m)=h^{(n)}-h^{(m)}$, where $h^{(n)}$ stores a ranking history of the $n$-th sample in a training dataset. We assume that a network prediction is correct~(\ie,~$\hat{y}=y$) for LS~\cite{szegedy2016rethinking}.}
   \vspace{-1mm}
   \label{tab:grad}
   \centering
   \renewcommand{\arraystretch}{1.5}
   \setlength{\tabcolsep}{2.2pt} 
   \resizebox{\linewidth}{!}{%
   \small
   \begin{tabular}{c|l|C{4cm}C{4cm}|C{4cm}C{4cm}}
   \hline
   \multirow{2}{*}{\rb{Type}} &
     \multicolumn{1}{c|}{\multirow{2}{*}{Method}} &
     \multicolumn{2}{c|}{Smoothing function $f$} &
     \multicolumn{2}{c}{Indicator function $\mathbbm{C}$} \\ \cline{3-6} 
    &
     \multicolumn{1}{c|}{} &
     \multicolumn{1}{c|}{$j = \hat{y}$} &
     $ j \ne \hat{y}$ &
     \multicolumn{1}{c|}{$j = \hat{y}$} &
     $j \ne \hat{y}$ \\ \hline\hline
   \multirow{2}{*}{\rb{-}} &
     LS~\cite{szegedy2016rethinking} &
     \multicolumn{1}{c|}{$\epsilon_{1}$} &
     $\epsilon_{2}$ &
     \multicolumn{1}{c|}{$1$} &
     $1$ \\
     &
     FLSD~\cite{mukhoti2020calibrating} &
     \multicolumn{1}{c|}{$\lambda_{1}$} &
     $\lambda_{2}$ &
     \multicolumn{1}{c|}{$1$} &
     $1$ \\ \hline
   \multirow{2}{*}{\rb{AR}} &
     CPC~\cite{cheng2022calibrating} &
     \multicolumn{1}{c|}{$-\lambda_{1} \sum_{k \ne j}^{C} \frac{{p}_{k}}{{p}_{k} + {p}_{j}}$} &
     $\lambda_{1} \frac{{p}_{j}}{{p}_{y} + {p}_{j}} + \lambda_{2} \sum_{k \ne y} \frac{{p}_{k} - {p}_{j}}{{p}_{k} + {p}_{j}}$ &
     \multicolumn{1}{c|}{$1$} &
     $1$ \\
     &
     MDCA~\cite{hebbalaguppe2022stitch} &
     \multicolumn{1}{c|}{$\lambda_{1} {p}_{j} (1 - {p}_{j})$} &
     $\lambda_{2} {p}_{j} (1 + {p}_{\hat{y}})$ &
     \multicolumn{1}{c|}{$1$} &
     $1$ \\ \hline
   \rb{CR} &
     MbLS~\cite{liu2022devil} &
     \multicolumn{1}{c|}{$\lambda_{1}$} &
     $\lambda_{2}$ &
     \multicolumn{1}{c|}{$0$} &
     $\mathbbm{1}[{z}_{\hat{y}} - {z}_{j} \ge M]$ \\ \hline
   \multirow{2}{*}{\rb{ACR}} &
     CRL~\cite{moon2020confidence} &
     \multicolumn{1}{c|}{$\lambda_{1} {p}_{j}(1 - {p}_{j})$} &
     $\lambda_{2} {p}_{\hat{y}}{p}_{j}$ &
     \multicolumn{1}{c|}{$\mathbbm{1}[ H(n, m) ({p}_{\hat{y}}^{(n)} - {p}_{\hat{y}}^{m} ) < 0]$} &
     $\mathbbm{1}[ H(n, m) ({p}_{\hat{y}}^{(n)} - {p}_{\hat{y}}^{m} ) < 0]$ \\ 
   &
     Ours~(ACLS) &
     \multicolumn{1}{c|}{$\lambda_{1} \left( z_{j} -\min_{k} z_{k} - M \right)$} &
     $\lambda_{2} \left( z_{\hat{y}} - z_{j} -M \right)$ &
     \multicolumn{1}{c|}{$\mathbbm{1}[{z}_{j} - \min_{k} {z}_{k} \ge M]$} &
     $\mathbbm{1}[{z}_{\hat{y}} - {z}_{j} \ge M]$ \\ \hline
   \end{tabular}
   }
   \vspace{-3mm}
\end{table*}

In this section, we first describe preliminaries on network calibration~(Sec.~\ref{sec:3.1}). We then provide an in-depth analysis of regularization-based methods~\cite{cheng2022calibrating,hebbalaguppe2022stitch,liu2022devil,moon2020confidence,lin2017focal,mukhoti2020calibrating} in terms of gradients of objective functions~(Sec.~\ref{sec:3.2}), and introduce our loss function, ACLS, for network calibration~(Sec.~\ref{sec:3.3}).

\subsection{Network calibration} \label{sec:3.1}
Given an input $\mathbf{x}$ and a corresponding ground-truth label~(class) $y$, DNNs output a logit vector $\mathbf{z} \in \mathbb{R}^{C}$, where $C$ is the number of classes. To train DNNs, a softmax CE loss is generally used as follows:
\begin{equation} \label{eq:CE}
   \mathcal{L}_{\text{CE}} =  \mathbb{E}_{\mathbf{q}} \left[ \mathbf{p} \right] = - \sum_{k=1}^{C} q_{k} \log p_{k},
\end{equation}
where $\mathbf{p}$ represents a softmax output of the network, and $\mathbf{q}$ indicates a target distribution~(\ie, a distribution of one-hot encoded labels). We denote by $p_{j}$ and $q_{j}$ elements of $\mathbf{p}$ and $\mathbf{q}$ for the class~$j$, respectively. We define the softmax output for the class~$j$ as follows: 
\begin{equation} \label{eq:softmax}
   p_{j} =  \frac{\exp (z_{j})}{\sum_{k=1}^{C} \exp (z_{k})},
\end{equation}
where we denote by $z_{j}$ the logit value of $\mathbf{z}$ for the class~$j$.
Formally, we compute gradients of $\mathcal{L}_{\text{CE}}$ w.r.t the logit for the class $j$ as follows:
\begin{equation}\label{eq:ce_gradient}
   \frac{\partial \mathcal{L}_{\text{CE}}}{\partial {z}_{j}} = {p}_{j} - {q}_{j}
\end{equation}
The CE loss encourages minimizing the discrepancies between probability distributions of $\mathbf{p}$ and $\mathbf{q}$, learning discriminative feature representations. To this end, the probability~$p_{y}$ for the true class~$y$ should be one, the same as $q_y$. Accordingly, the CE loss tries to raise the corresponding logit~$z_y$~(or equivalently the probability~$p_{y}$), although the probability~$p_{y}$ could not reach to the value of one exactly~\cite{szegedy2016rethinking}, resulting in the overconfidence problem~\cite{guo2017calibration,mukhoti2020calibrating,cheng2022calibrating,liu2022devil,szegedy2016rethinking}. 

\vspace{-2mm}
\paragraph{Label smoothing.} To address the overconfidence problem, label smoothing~\cite{szegedy2016rethinking} modifies the target distribution as follows:
\begin{equation} \label{eq:ls_labels}
   {q}^{\prime}_{j} =
   \begin{cases}
      {q}_{j} - \epsilon \left( 1 - \frac{1}{C} \right),       & j = y\\
      {q}_{j} + \frac{\epsilon}{C},                            & j \ne y
   \end{cases},
\end{equation}
where $\epsilon$ is a hyperparameter for controlling the degree of smoothing. It lowers the initial target probability~$q_{y}$ for the true class~$y$, while raising the probabilities for others. For better understanding its behavior, we compute the gradients of the objective function for label smoothing~$\mathcal{L}_{\text{LS}}$ w.r.t the logit~$z_{j}$, which is the same as the CE loss in Eq.~\eqref{eq:CE} but with the target distribution of~${q}^{\prime}_{j}$, as follows:
\begin{equation}\label{eq:ls_gradient}
   \frac{\partial \mathcal{L}_{\text{LS}}}{\partial {z}_{j}} = {p}_{j} - {q}^{\prime}_{j}=
   \begin{cases}
      {p}_{j} - \left({q}_{j} - \epsilon_{1} \right),       & j = y\\
      {p}_{j} - \left({q}_{j} + \epsilon_{2} \right),       & j \ne y
   \end{cases},
\end{equation}
where $\epsilon_{1} = \epsilon ( 1 - 1/C )$ and $\epsilon_{2} = {\epsilon}/{C}$. In contrast to the CE, the probability of $p_{y}$ for the true class~$y$ would be the same as that of $q_{y} - \epsilon_{1}$. The logit value~$z_y$ will thus not be raised continually, mitigating the overconfidence problem. Note that the gradients for label smoothing in Eq.~\eqref{eq:ls_gradient} reduce to those for CE in Eq.~\eqref{eq:ce_gradient} with $\epsilon$ being zero. In the following, we will show that existing regularization-based methods can be considered as variants of label smoothing.

\subsection{Gradient analysis} \label{sec:3.2}
Regularization approaches to network calibration generally exploit a regularization term along with a fidelity being the CE loss to alleviate the miscalibration problem. Although they have proven effective in network calibration, there is a lack of theoretical understandings on how the regularization term affects on the calibration. To better understand the effects on calibration, we compute gradients of objective functions for existing regularization-based methods~(Table~\ref{tab:grad}). 

Concretely, we can represent the loss functions of calibration methods~\cite{cheng2022calibrating,hebbalaguppe2022stitch,liu2022devil,moon2020confidence,lin2017focal,mukhoti2020calibrating} as follows:
\begin{equation} \label{eq:loss}
   \mathcal{L} = \mathcal{L}_{\text{CE}} + \mathcal{L}_{\text{REG}},
\end{equation}
where $\mathcal{L}_{\text{REG}}$ is a regularization term.
We compute the gradient of Eq.~\eqref{eq:loss} w.r.t the logit value $z_{j}$ for the class~$j$:
\begin{equation} \label{eq:grad}
   \frac{\partial \mathcal{L}}{\partial z_{j}} =  p_{j} - q_{j} + \frac{\partial \mathcal{L}_{\text{REG}}}{\partial z_{j}},
\end{equation}
which can be reformulated as follows:
\begin{equation} \label{eq:grad_unified}
   \frac{\partial \mathcal{L}}{\partial {z}_{j}} =
   \begin{cases}
      {p}_{j} - \left({q}_{j} - f({z}_{j})\mathbbm{C}({z}_{j}) \right), & j = \hat{y} \\ 
      {p}_{j} - \left({q}_{j} + f({z}_{j})\mathbbm{C}({z}_{j}) \right), & j \ne \hat{y}
   \end{cases},
\end{equation}
where we denote by $\hat{y}$ a network prediction. $f$ is a smoothing function that determines the degree of smoothing, and $\mathbbm{C}$ is an indicator function deciding whether to apply smoothing or not. Note that the network calibration methods are designed to lower the largest output probability~$p_{\hat{y}}$ across the classes, where the overconfidence problem is likely to occur. The gradients in Eq.~\eqref{eq:grad_unified} are thus split based on the network prediction~$\hat{y}$, in contrast to label smoothing~\cite{szegedy2016rethinking} using the ground-truth label~$y$ in Eq.~\eqref{eq:ls_gradient}. We can see that 1) the gradients of existing regularization-based methods in Eq.~\eqref{eq:grad_unified} generalize those for label smoothing in Eq.~\eqref{eq:ls_gradient} if the prediction is correct~(\ie,~$\hat{y} = y$), and 2) each method differs in how smoothing and indicator functions are defined~(Table~\ref{tab:grad}), suggesting that the regularization-based methods can be regarded as a form of label smoothing. We provide detailed derivations and explanations of smoothing and indicator functions for each method in the supplementary material. 

\begin{figure}[]
   \begin{center}
   \begin{subfigure}{0.47\columnwidth}
   \includegraphics[width=\textwidth]{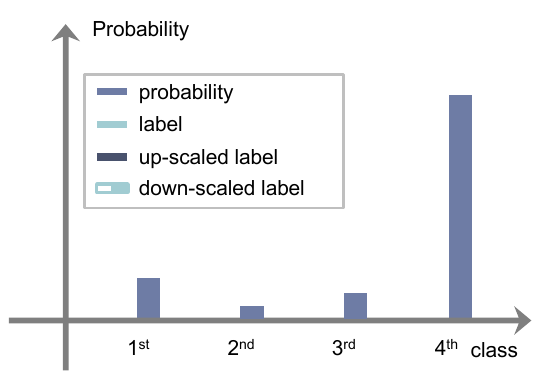}
   \caption{Probabilities~($\mathbf{p}$).}
   \end{subfigure}
   \begin{subfigure}{0.47\columnwidth}
   \includegraphics[width=\textwidth]{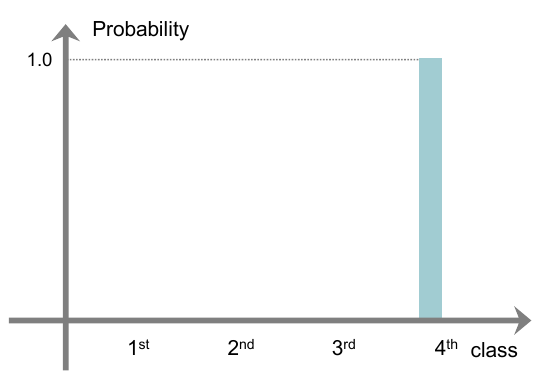}
   \caption{Labels ($\mathbf{q}$).}

   \end{subfigure}
   \begin{subfigure}{0.47\columnwidth}
   \includegraphics[width=\textwidth]{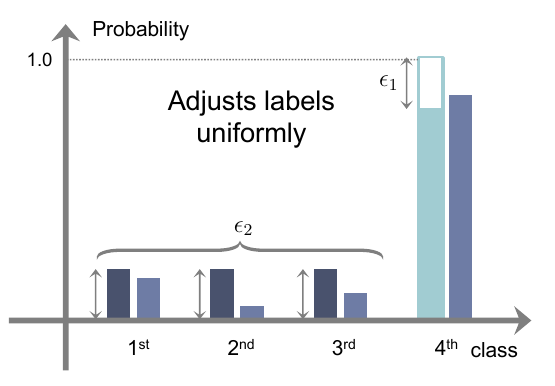}
   \caption{LS~\cite{szegedy2016rethinking}.}
   \end{subfigure} 
   \begin{subfigure}{0.47\columnwidth} 
   \includegraphics[width=\textwidth]{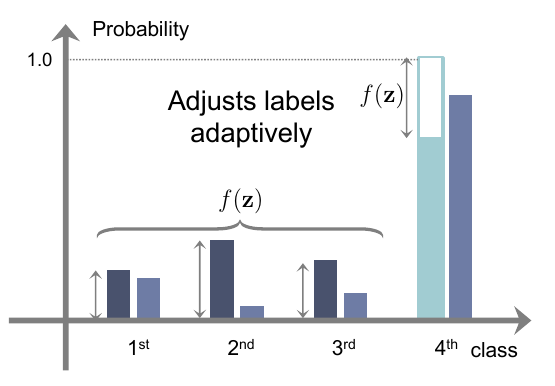} 
   \caption{AR~\cite{cheng2022calibrating,hebbalaguppe2022stitch}.}
   \end{subfigure}  

   \begin{subfigure}{0.47\columnwidth} 
   \includegraphics[width=\textwidth]{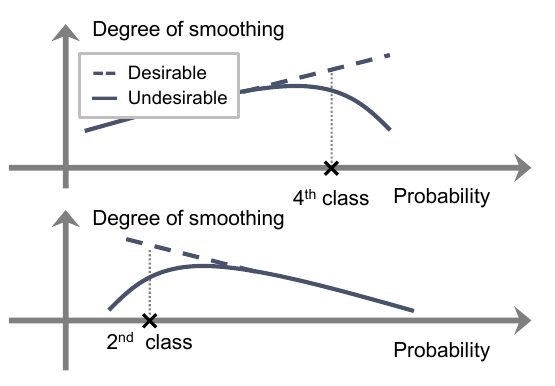} 
   \caption{\centering Undesirable smoothing behaviors of AR.}
   \end{subfigure}
   \begin{subfigure}{0.47\columnwidth} 
   \includegraphics[width=\textwidth]{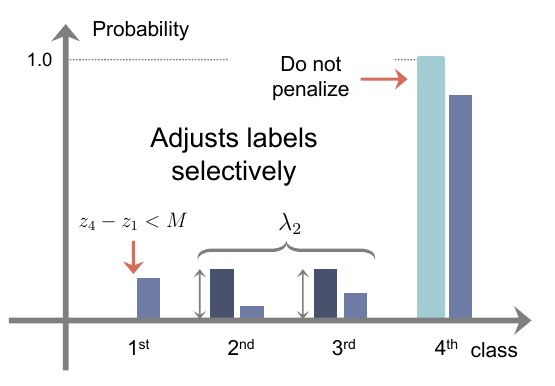} 
   \caption{CR~\cite{liu2022devil}.\\ ~} 
   \end{subfigure}
   \end{center}
   \vspace{-4mm}
   \captionsetup{font={small}}
   \caption{Illustration of smoothing behaviors of regularization-based approaches to network calibration: (c)~LS~\cite{szegedy2016rethinking}, (d)~AR~\cite{cheng2022calibrating,hebbalaguppe2022stitch}, and (f)~CR~\cite{liu2022devil}. We visualize the case that the network prediction is correct~(\ie,~$\hat{y}=y$), but overconfident, with (a)~softmax probabilities and (b)~target labels. We also illustrate the undesirable smoothing behaviors of AR in (e). Best viewed in color.}
   \vspace{-4mm}
   \label{fig:prob}
\end{figure}

According to $f$ and $\mathbbm{C}$, we can divide existing methods into three groups: AR, CR, and a combination of them~(ACR). We illustrate in Fig.~\ref{fig:prob} the behavior of each regularization method on calibrating output probabilities. Given softmax probabilities~$\mathbf{p}$ and a target distribution~$\mathbf{q}$~(\ie,~training labels), LS~\cite{szegedy2016rethinking} uniformly adjusts the distribution with hyperparameters,~\ie, $\epsilon_1$ and $\epsilon_2$~(Fig.~\ref{fig:prob}(c)). That is, it uses a constant smoothing function without an indicator function (See the first row in Table~\ref{tab:grad}). FLSD~\cite{mukhoti2020calibrating}, which is a variant of LS, also adopts a uniform smoothing function~(See the second row in Table~\ref{tab:grad}).

\vspace{-4mm}
\paragraph{AR.} 
 In contrast to LS~\cite{szegedy2016rethinking}, AR methods use an adaptive smoothing function. We can see in Fig.~\ref{fig:prob}(d) that they adjust the degree of smoothing non-uniformly according to logit values. For example, the smoothing function increases in proportion to the logits, if $j=\hat{y}$, reducing the target labels of ${q}_{j} - f({z}_{j})$ in Eq.~\eqref{eq:grad_unified}~(\eg,~the fourth class in Fig.~\ref{fig:prob}(d)). For $j \ne \hat{y}$, the smoothing function raises the target labels of~${q}_{j} + f({z}_{j})$ in Eq.~\eqref{eq:grad_unified} in accordance with the logits~(\eg,~from the first to third classes in Fig.~\ref{fig:prob}(d)). In this case, the probabilities for the corresponding classes increase, making the probability of a network prediction, $p_{\hat{y}}$, relatively small. This reduces the overconfident probability (\eg, $p_{4}$ in Fig.~\ref{fig:prob}(d)) more effectively than LS. However, we have observed that the smoothing function of AR often violates the desirable behaviors. For example, the smoothing function of MDCA~\cite{hebbalaguppe2022stitch} has a parabolic form, which is problematic. When the logit (or the softmax probability of~$p_{j}$) is extremely large for $j = \hat{y}$~(the top in Fig.~\ref{fig:prob}(e)), MDCA rather lessens the degree of smoothing, even worsening the overconfidence problem. Similarly, it lessens the degree of smoothing, when the logit is very small for $j \ne \hat{y}$~(the bottom in Fig.~\ref{fig:prob}(e)). CPC~\cite{cheng2022calibrating} outputs negative values when $j=\hat{y}$. This in turn raises the target label of the overconfident probability, lessening the degree of smoothing and disturbing a calibration process.

\vspace{-4mm}
\paragraph{CR.} Instead of using adaptive smoothing functions, a CR method~\cite{liu2022devil} adopts a constant function as in LS~\cite{szegedy2016rethinking}, and performs label smoothing for miscalibrated probabilities only. In particular, it exploits a specific condition as follows:
\begin{equation} \label{eq:ex_mbls}
   \mathbbm{C}({z}_{j}) =
   \begin{cases}
      0,                                                                & j = \hat{y} \\
      \mathbbm{1}[{z}_{\hat{y}} - {z}_{j} \ge M],                       & j \ne \hat{y}
   \end{cases},
\end{equation}
where $M$ is a margin and we denote by $\mathbbm{1}[\cdot]$ a function whose output is 1 when the argument is true, and $0$ otherwise. We can see that the CR method~\cite{liu2022devil} raises the target labels of ${q}_{j} + f({z}_{j})$ in Eq.~\eqref{eq:grad_unified} only if $z_{\hat{y}} - z_{j}$ is larger than the margin $M$ for $j \ne \hat{y}$. In other words, it determines that the probabilities~$p_j$, where $j \ne \hat{y}$, are well-calibrated, if the logit difference between the network prediction and other classes is smaller than the margin~$M$, and applies regularization for miscalibrated probabilities selectively. For example, the first class in Fig.~\ref{fig:prob}(f) does not satisfy the condition in Eq.~\eqref{eq:ex_mbls}, and thus regularization is not applied, while label smoothing is employed by the amount of~$\lambda_{2}$ for the second and third classes~(See Table~\ref{tab:grad}). However, the CR method~\cite{liu2022devil} does not penalize the target label for the class $\hat{y}$ (\ie,~$\mathbbm{C}({z}_{j})=0$ for $j = \hat{y}$) directly. This degrades the calibration performance, since the overconfidence problem is likely to occur in the largest probability of $p_{\hat{y}}$~(\eg,~the fourth class in Fig~\ref{fig:prob}(f)). Note that applying the softmax across $p_{j}$ for all $j$ in the CR method~\cite{liu2022devil} could lower the probability of $p_{\hat{y}}$ for the class $\hat{y}$ implicitly. However, by reducing the target label of the class $\hat{y}$ explicitly, while raising the labels of other classes $j \ne \hat{y}$, similar to AR and ACR, we can lower the overconfident probability of $p_{\hat{y}}$ more effectively, mitigating the overconfidence problem.

\vspace{-4mm}
\paragraph{ACR.} CRL~\cite{moon2020confidence} has the merits of both AR and CR, adaptively penalizing target labels and preserving well-calibrated probabilities. However, it also inherits the drawback of AR. For example, the smoothing function of CRL~\cite{moon2020confidence} has a parabolic form, when $j=\hat{y}$ and $\mathbbm{C}=1$, which is the same as the function of MDCA~\cite{hebbalaguppe2022stitch}. This violates the desirable behaviors of the smoothing function~(See the top in Fig.~\ref{fig:prob}(e)). Note that the indicator function of CRL uses an ordinal ranking condition~\cite{corbiere2019addressing}. The ranking criterion first records how many times each sample is classified correctly during training~(\ie, correctness). It then compares the ordinal ranking relationship based on the correctness by exploiting a ranking loss~\cite{wang2014learning}. At the early phase of training, the correctness history is empty. Networks are thus trained using the CE loss only, causing the overconfidence problem. Moreover, the correctness increases, as the training process goes on. The ranking condition of the indicator function is thus easily not satisfied, making the regularizer inactive and degrading the calibration performance~(See Sec~\ref{sec:4.3}).

\subsection{ACLS} \label{sec:3.3}
We propose novel smoothing and indicator functions that avoid the negative effects of AR and CR, while retaining the advantages of both approaches.

\vspace{-4mm}
\paragraph{Smoothing function.}
To address the limitation of AR~\cite{hebbalaguppe2022stitch,cheng2022calibrating}, we set a smoothing function:
\begin{equation} \label{eq:ours_smoothing}
   f({z}_{j}) =
   \begin{cases}
      {\lambda_{1}} \left( z_{j} -\min_{k} z_{k} - M \right),               & j = \hat{y} \\
      {\lambda_{2}} \left( z_{\hat{y}} - z_{j} -M \right),                  & j \ne \hat{y}
   \end{cases},
\end{equation}
where $\lambda_{1}$ and $\lambda_{2}$ are hyperparameters for $j= \hat{y}$ and $j \ne \hat{y}$ terms, respectively. For $j = \hat{y}$, our smoothing function provides outputs proportional to the logit values~$z_{j}$, as it is piecewise linear w.r.t $z_{j}$. This indicates that our target label in Eq.~\eqref{eq:grad_unified},~\ie, $q_{j} - f({z}_{j})$, decreases with an increment of the logit~$z_{j}$, thereby lowering the probability $p_{\hat{y}}$. Similarly, if $j \ne \hat{y}$, the smoothing function outputs larger values, in accordance with the decrement in the logit $z_{j}$, which in turn raises the target label,~\ie, $q_{j} + f({z}_{j})$, and the probability $p_j$. In summary, the piecewise linearity in our smoothing function enables always lowering the target label, as the probability $p_{\hat{y}}$ increases, and vice versa for other cases consistently, alleviating the limitation of AR. 
 
\vspace{-4mm}
\paragraph{Indicator function.}  
We design an indicator function of the gradients in Eq.~\eqref{eq:grad_unified} as follows:
\begin{equation} \label{eq:ours_decision}
   \mathbbm{C}({z}_{j}) =
   \begin{cases}
      \mathbbm{1}[{z}_{j} - \min_{k} {z}_{k} \ge M],                                 & j = \hat{y} \\
      \mathbbm{1}[{z}_{\hat{y}} - {z}_{j} \ge M],                       & j \ne \hat{y}
   \end{cases}.
\end{equation}
For $j\ne \hat{y}$, we adopt the margin criterion for MbLS~\cite{liu2022devil} to avoid the limitations of an ordinal ranking condition. Different from MbLS, we also use an indicator function for $j = \hat{y}$, enabling adjusting the target label for the class~$\hat{y}$, where the overconfidence problem is likely to occur. Specifically, when $j = \hat{y}$, the indicator function is $1$, if the difference between maximum and minimum logits is greater than the margin~$M$, addressing the overconfidence problem of MbLS.

\vspace{-4mm}
\paragraph{Training.} By putting our smoothing and indicator functions of Eq.~\eqref{eq:ours_smoothing} and Eq.~\eqref{eq:ours_decision}, respectively, into the gradient of a regularization term in Eq.~\eqref{eq:grad_unified}, we can obtain a novel form of gradients w.r.t the logit $z_{j}$ as follows:
\begin{equation} \label{eq:ours_grad}
   \frac{\partial \mathcal{L}_{\text{ACLS}}}{\partial {z}_{j}} =
   \begin{cases}
        \text{~~~} {\lambda_{1}} \text{ReLU} \left( z_{j} -\min_{k} z_{k} - M \right),                  & j = \hat{y} \\ 
      - {\lambda_{2}}\text{ReLU} \left( z_{\hat{y}} - z_{j} - M \right),                               & j \ne \hat{y}
   \end{cases}.
\end{equation}
By integrating the gradients, we obtain the loss function, ACLS, as follows: 
\begin{equation} \label{eq:ours_loss}
   \mathcal{L}_{\text{ACLS}} =
   \begin{cases}
      {\lambda_{1}} \left( \text{ReLU} \left( z_{j} -\min_{k} z_{k} - M \right) \right) ^{2},                           & j = \hat{y} \\ 
      {\lambda_{2}} \left( \text{ReLU} \left( z_{\hat{y}} - z_{j} - M \right) \right) ^{2},                             & j \ne \hat{y}
   \end{cases}.
\end{equation}

\noindent
Similar to other calibration methods~\cite{cheng2022calibrating,hebbalaguppe2022stitch,liu2022devil,moon2020confidence,lin2017focal,mukhoti2020calibrating}, we use both fidelity and regularization terms to train our model. An overall loss function is as follows:
\begin{equation} \label{eq:total_loss}
   \mathcal{L} = \mathcal{L}_{\text{CE}} + \mathcal{L}_{\text{ACLS}}.
\end{equation}

\vspace{-4mm}
\paragraph{ACLS for network calibration.} Regularizers for network calibration satisfy the following conditions:~(1)~They could lower a probability of $p_{\hat{y}}$ for a class $\hat{y}$,~\ie,~the largest probability across classes, where the overconfidence problem is likely to occur.~(2)~They should penalize the probability of $p_{\hat{y}}$ selectively, only when it is miscalibrated, since the overconfidence problem for $p_{\hat{y}}$ does not always occur. ACLS in Eq.~\eqref{eq:ours_loss} satisfies these conditions. Specifically, it can reduce a logit for the class $\hat{y}$, lowering the probability of $p_{\hat{y}}$ for the class $\hat{y}$, and vice versa for other classes. The $\text{ReLU}$ function with the margin $M$ determines whether to adjust the labels, regularizing logit values selectively.

\begin{table*}[]
   \centering
   \captionsetup{font={small}}
   \caption{Quantitative results on the validation split of CIFAR10~\cite{krizhevsky2009learning}, Tiny-ImageNet~\cite{le2015tiny}, and ImageNet~\cite{deng2009imagenet} in terms of the top-1 accuracy~(ACC), ECE, and AECE. We compute the calibration metrics with 15 bins. Numbers in bold are the best performance and underlined ones are the second best.}
   \vspace{-2mm}
   \label{tab:cls}
   \renewcommand{\arraystretch}{1.0}
   \setlength{\tabcolsep}{2.5pt} 
   \small
   \begin{tabular}{l|C{0.81cm}C{0.81cm}C{0.81cm}|C{0.81cm}C{0.81cm}C{0.81cm}|C{0.81cm}C{0.81cm}C{0.81cm}|C{0.81cm}C{0.81cm}C{0.81cm}|C{0.81cm}C{0.81cm}C{0.81cm}}
   \hline
   \multicolumn{1}{c|}{Dataset} & \multicolumn{6}{c|}{CIFAR10} & \multicolumn{6}{c|}{Tiny-ImageNet} & \multicolumn{3}{c}{ImageNet} \\ \hline \hline
   \multicolumn{1}{c|}{\multirow{2}{*}{\diagbox{\small{Method}}{\small{Arch.}}}} &
     \multicolumn{3}{c|}{ResNet-50} &
     \multicolumn{3}{c|}{ResNet-101} &
     \multicolumn{3}{c|}{ResNet-50} &
     \multicolumn{3}{c|}{ResNet-101} &
     \multicolumn{3}{c}{ResNet-50} \\ \cline{2-16} 
   \multicolumn{1}{c|}{} &
     \multicolumn{1}{c}{\scriptsize{~~ACC~$\uparrow$}} &
     \multicolumn{1}{c}{\scriptsize{~~ECE~$\downarrow$}} &
     \multicolumn{1}{c|}{\scriptsize{AECE$\downarrow$}} &
     \multicolumn{1}{c}{\scriptsize{~~ACC~$\uparrow$}} &
     \multicolumn{1}{c}{\scriptsize{~~ECE~$\downarrow$}} &
     \multicolumn{1}{c|}{\scriptsize{AECE$\downarrow$}} &
     \multicolumn{1}{c}{\scriptsize{~~ACC~$\uparrow$}} &
     \multicolumn{1}{c}{\scriptsize{~~ECE~$\downarrow$}} &
     \multicolumn{1}{c|}{\scriptsize{AECE$\downarrow$}} &
     \multicolumn{1}{c}{\scriptsize{~~ACC~$\uparrow$}} &
     \multicolumn{1}{c}{\scriptsize{~~ECE~$\downarrow$}} &
     \multicolumn{1}{c|}{\scriptsize{AECE$\downarrow$}} &
     \multicolumn{1}{c}{\scriptsize{~~ACC~$\uparrow$}} &
     \multicolumn{1}{c}{\scriptsize{~~ECE~$\downarrow$}} &
     \multicolumn{1}{c}{\scriptsize{AECE$\downarrow$}} \\ \hline \hline
   CE &
     \multicolumn{1}{c}{93.20} &
     \multicolumn{1}{c}{5.85} &
     \multicolumn{1}{c|}{5.84} &
     \multicolumn{1}{c}{93.33} &
     \multicolumn{1}{c}{5.74} &
     \multicolumn{1}{c|}{5.73} &
     \multicolumn{1}{c}{65.02} &
     \multicolumn{1}{c}{3.73} &
     \multicolumn{1}{c|}{3.69} &
     \multicolumn{1}{c}{65.62} &
     \multicolumn{1}{c}{4.97} &
     \multicolumn{1}{c|}{4.97} &
     \multicolumn{1}{c}{73.96} &
     \multicolumn{1}{c}{9.10} &
     \multicolumn{1}{c}{9.24} \\
   CE+TS\fm~\cite{guo2017calibration} & 93.20             & 3.68             & 3.67             & 93.33             & 3.62             & 3.62             & 65.02             & 1.63             & 1.52             & 65.62             & 2.08             & 2.03             & 73.96             & 1.86               & 1.90             \\
   MMCE~\cite{kumar2018trainable}     & 95.18             & 3.10             & 3.10             & 94.99             & 3.61             & \underline{3.61} & 64.75             & 5.15             & 5.12             & 65.92             & 4.88             & 4.88             & 74.36             & 8.75               & 8.75             \\
   ECP~\cite{pereyra2017regularizing} & 94.75             & 3.01             & \underline{2.99} & 93.35             & 5.41             & 5.40             & 64.98             & 4.00             & 3.92             & 65.69             & 4.68             & 4.66             & 73.84             & 8.63               & 8.63             \\
   LS~\cite{szegedy2016rethinking}    & 94.87             & 2.79             & 3.85             & 93.23             & 3.56             & 4.68             & \bf{65.78}        & 3.17             & 3.16             & 65.87             & 2.20             & 2.21             & 75.24             & 2.33               & 2.43             \\
   FL~\cite{lin2017focal}             & 94.82             & 3.90             & 3.86             & 92.42             & 4.60             & 4.58             & 63.09             & 2.96             & 3.12             & 62.97             & 2.55             & 2.44             & 71.76             & \underline{1.79}  & 1.79             \\
   FLSD~\cite{mukhoti2020calibrating} & 94.77             & 3.84             & 3.60             & 92.38             & 4.58             & 4.57             & 64.09             & 2.91             & 2.95             & 62.96             & 4.91             & 4.91             & 72.19             & 1.82               & \underline{1.86} \\
   MDCA~\cite{hebbalaguppe2022stitch} & 94.84             & 6.86             & 6.73             & \bf{95.66}        & 6.88             & 7.23             & 63.57             & 2.77             & 2.61             & \underline{66.34} & 6.06             & 6.06             & \underline{75.65} & 7.45               & 7.53             \\
   CPC~\cite{cheng2022calibrating}    & 95.04             & 3.91             & 3.91             & \underline{95.36} & 3.78             & 3.75             & \underline{65.44} & 3.12             & 3.05             & \bf{66.56}        & 3.90             & 4.00             & \bf{75.76}        & 4.80               & 4.82             \\
   MbLS~\cite{liu2022devil}           & \underline{95.25} & \underline{1.16} & 3.18             & 95.13             & \underline{1.38} & \bf{3.25}        & 64.74             & \underline{1.64} & 1.73             & 65.81             & \underline{1.62} & \underline{1.68} & 75.39             & 4.07               & 4.14             \\ 
   CRL~\cite{moon2020confidence}      & 95.08             & 3.14             & 3.11             & 95.04             & 3.74             & 3.73             & 64.88             & 1.65             & \underline{1.52} & 65.87             & 3.57             & 3.56             & 73.83             & 8.47               & 8.47             \\ \hline
   ACLS                               & \bf{95.40}        & \bf{1.12}        & \bf{2.87}        & 95.34             & \bf{1.36}        & \bf{3.25}        & 64.84             & \bf{1.05}        & \bf{1.03}        & 65.78             & \bf{1.11}        & \bf{1.15}        & \underline{75.65} & \bf{1.02}              & \bf{1.20}        \\ \hline
   \end{tabular}
   \vspace{-3mm}
\end{table*}

\section{Experiments}
In this section, we describe implementation details (Sec~\ref{sec:4.1}) and compare our approach with the state of the art on image classification and semantic segmentation in terms of calibration metrics~(Sec.~\ref{sec:4.2}). We also provide a detailed analysis of our loss function~(Sec~\ref{sec:4.3}). 

\subsection{Implementation details.} \label{sec:4.1}

\vspace{-1mm}
\paragraph{Datasets and evaluation.}
We evaluate our method on CIFAR10~\cite{krizhevsky2009learning}, Tiny-ImageNet~\cite{le2015tiny}, ImageNet~\cite{deng2009imagenet} for image classification, and PASCAL VOC~\cite{everingham2009pascal} for semantic segmentation. The CIFAR10 dataset consists of 50K training and 10K test images of size $32 \times 32$ for 10 classes. Tiny-ImageNet is a subset of ImageNet, and provides 100K training, 10K validation, and 10K test images of 200 classes, where all images are downsampled to the size of $64 \times 64$ from ImageNet. The ImageNet dataset provides approximately 1.2M training and 50K validation images of 1K classes. PASCAL VOC presents 10,582 training and 1,449 validation samples. Following the standard protocol in~\cite{guo2017calibration,mukhoti2020calibrating,liu2022devil,hebbalaguppe2022stitch,moon2020confidence,minderer2021revisiting}, we report expected calibration error~(ECE~(\%)), adaptive ECE~(AECE~(\%)), and top-1 accuracy~(ACC~(\%)) for image classification. For semantic segmentation, we report ECE~(\%), AECE~(\%), and mIoU~(\%).

\vspace{-4mm}
\paragraph{Training.}
We use network architectures of ResNet-50 and ResNet-101~\cite{he2016deep} on CIFAR10~\cite{krizhevsky2009learning} and Tiny-ImageNet~\cite{le2015tiny}. For these datasets, we train the networks using the SGD optimizer with learning rate, weight decay, and momentum of 0.1, 5e-4, and 0.9, respectively. The networks are trained for 350 and 100 epochs with a batch size of 128 and 64 for CIFAR10 and Tiny-ImageNet, respectively. We divide the learning rate by 10 at 150 and 250 epochs for CIFAR10 and at 40 and 60 epochs for Tiny-ImageNet. We train our models with a single NVIDIA RTX A5000 GPU. For ImageNet~\cite{deng2009imagenet}, we train ResNet-50 for 200 epochs using the AdamW optimizer~\cite{loshchilov2017decoupled}, with batch size, learning rate, weight decay, $\beta_{1}$, and $\beta_{2}$ of 256, 5e-4, 5e-2, 0.9, and 0.999, respectively. We use a cosine annealing technique~\cite{loshchilov2016sgdr} as a learning rate scheduler, and train the network with four NVIDIA RTX A5000 GPUs. For semantic segmentation, we follow the experimental protocol in MbLS~\cite{liu2022devil}. We use DeepLabV3~\cite{chen2017rethinking} with ImageNet pretrained ResNet-50 as a backbone and train the model with a single NVIDIA RTX A5000 GPU. We train the entire model using the SGD optimizer for 100 epochs with batch size, learning rate, momentum, and weight decay of 8, 1e-2, 0.9, and 5e-4, respectively. We divide the learning rate by 10 at 40 and 80 epochs. 

\vspace{-4mm}
\paragraph{Hyperparameters.}
We set the margin~$M$ to 6 for CIFAR10, and 10 for other datasets, following~\cite{liu2022devil}. To set the hyperparameters $\lambda_{1}$ and $\lambda_{2}$ in Eq.~\eqref{eq:ours_smoothing}, we first divide 10\% of the training samples for each dataset as a cross-validation split. We then perform a grid search on the split, and set the values of 0.1 and 0.01 for $\lambda_{1}$ and $\lambda_{2}$, respectively. For other regularization-based methods~\cite{cheng2022calibrating,mukhoti2020calibrating,szegedy2016rethinking,hebbalaguppe2022stitch,liu2022devil,moon2020confidence}, we use default hyperparameters, provided by the authors. We provide a detailed analysis on these parameters in the supplementary material.

\footnotetext{We use the temperature scaling~(TS) method~\cite{guo2017calibration}, where temperature is optimized on the held-out validation splits. In our experiments, the optimal temperatures are 2.9, 1.1, and 1.4 for ResNet-50 on CIFAR10, Tiny-ImageNet, and ImageNet, and 2.9 and 1.1 for ResNet-101 on CIFAR-10 and Tiny-ImageNet, respectively.}

\begin{figure*}[]
   \begin{center}
       \begin{subfigure}{0.291\columnwidth}
           \includegraphics[width=\textwidth]{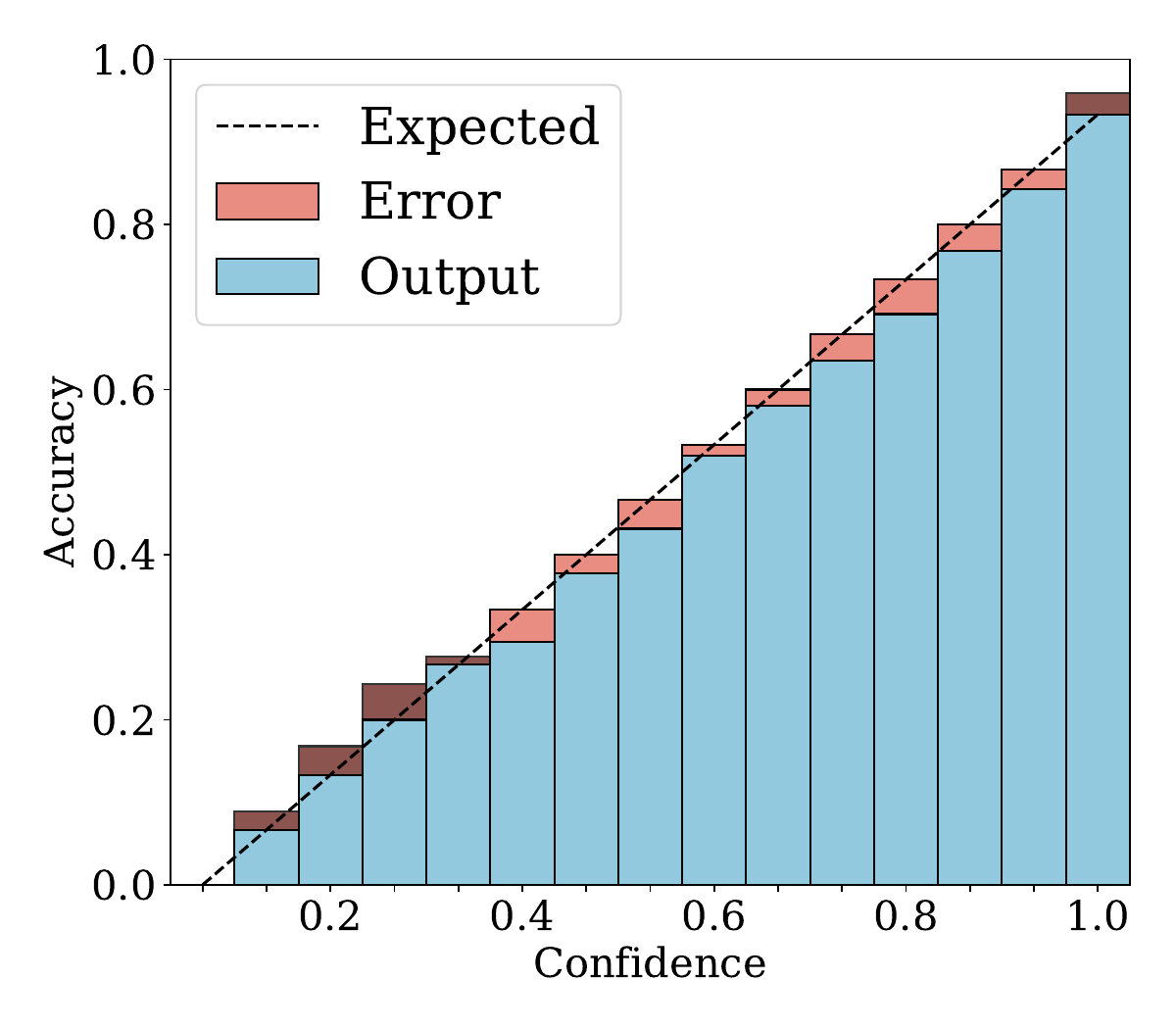}
       \end{subfigure}
       \begin{subfigure}{0.291\columnwidth}
           \includegraphics[width=\textwidth]{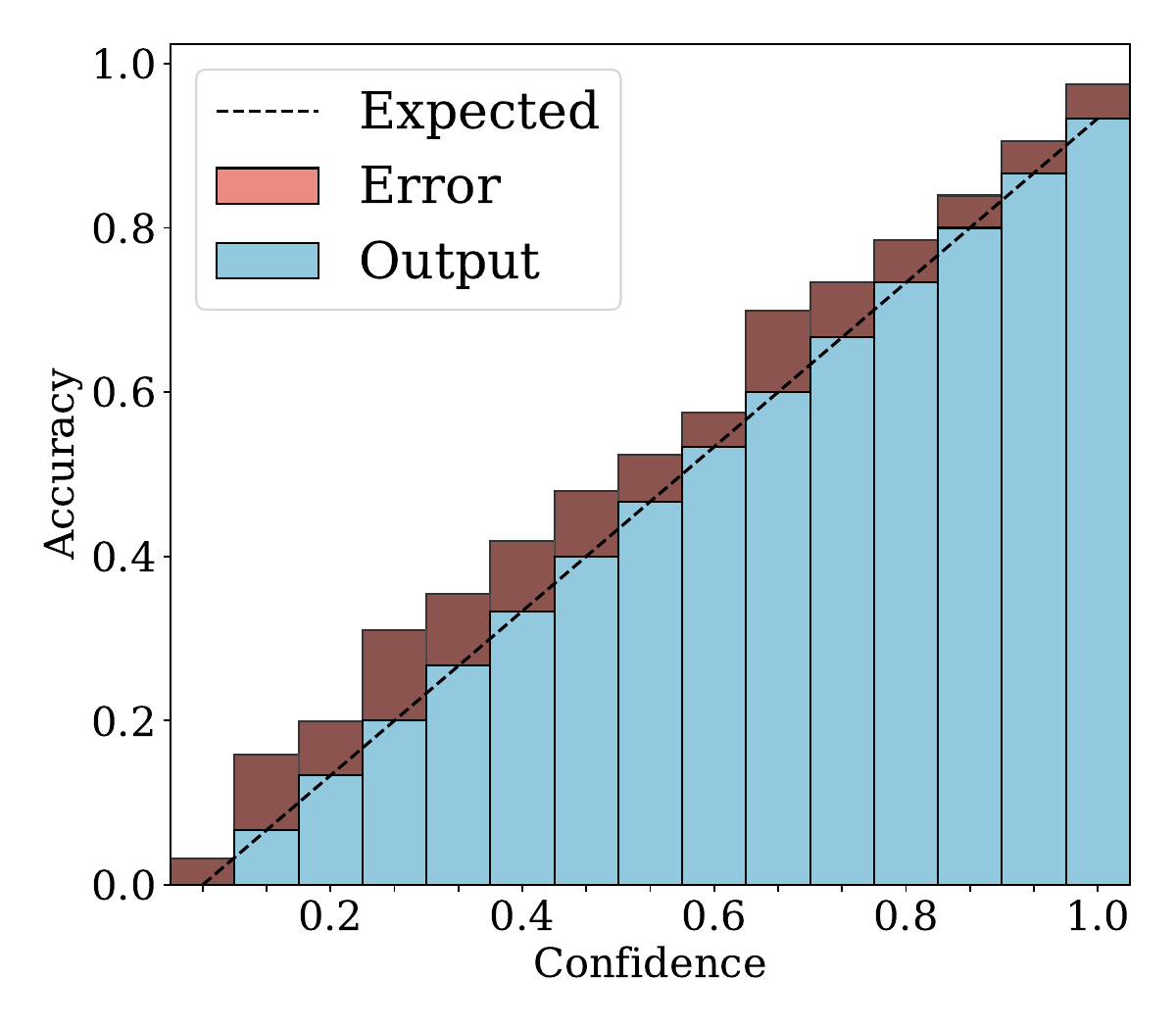}
       \end{subfigure} 
       \begin{subfigure}{0.291\columnwidth} 
           \includegraphics[width=\textwidth]{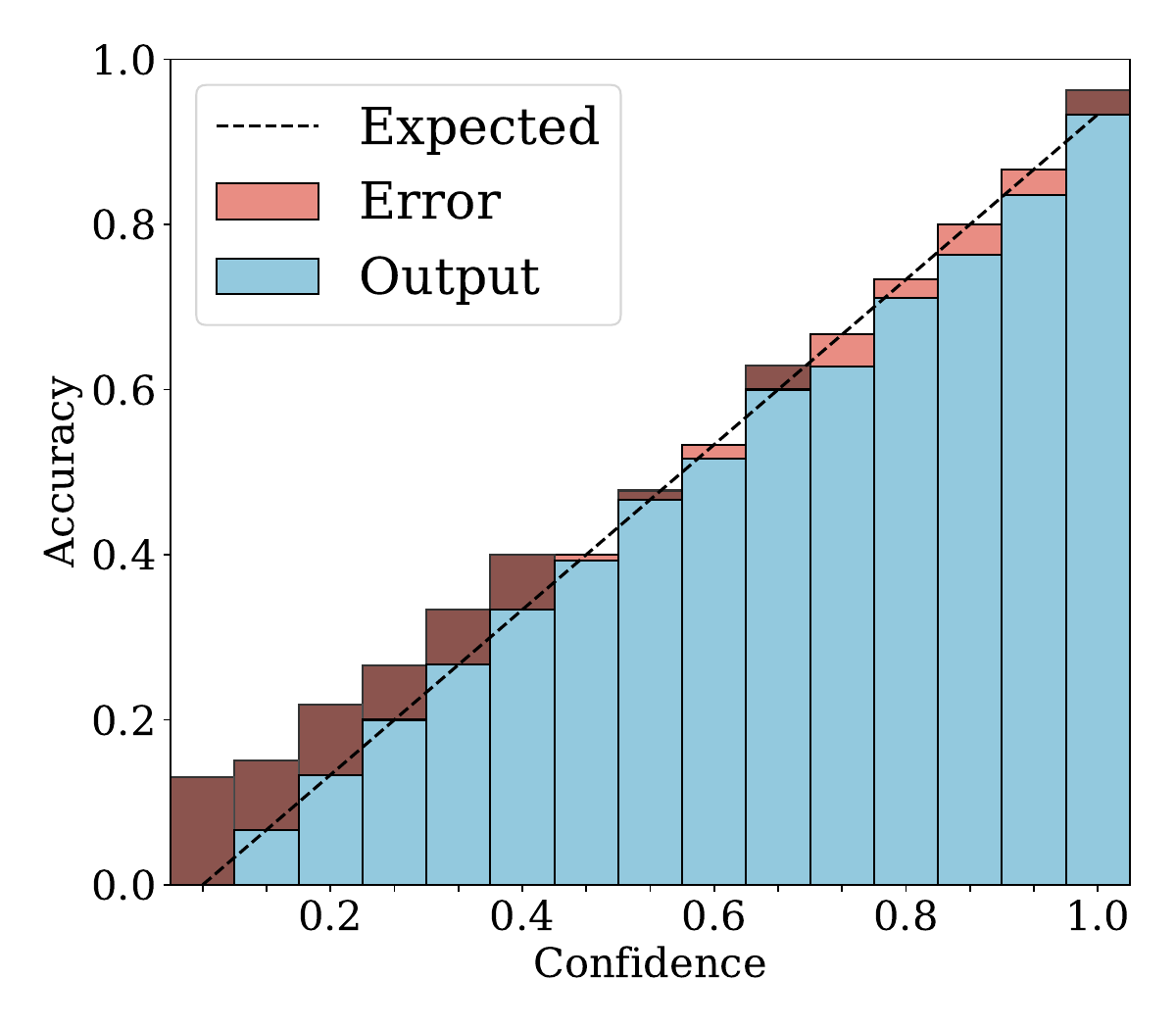} 
       \end{subfigure}
       \begin{subfigure}{0.291\columnwidth} 
           \includegraphics[width=\textwidth]{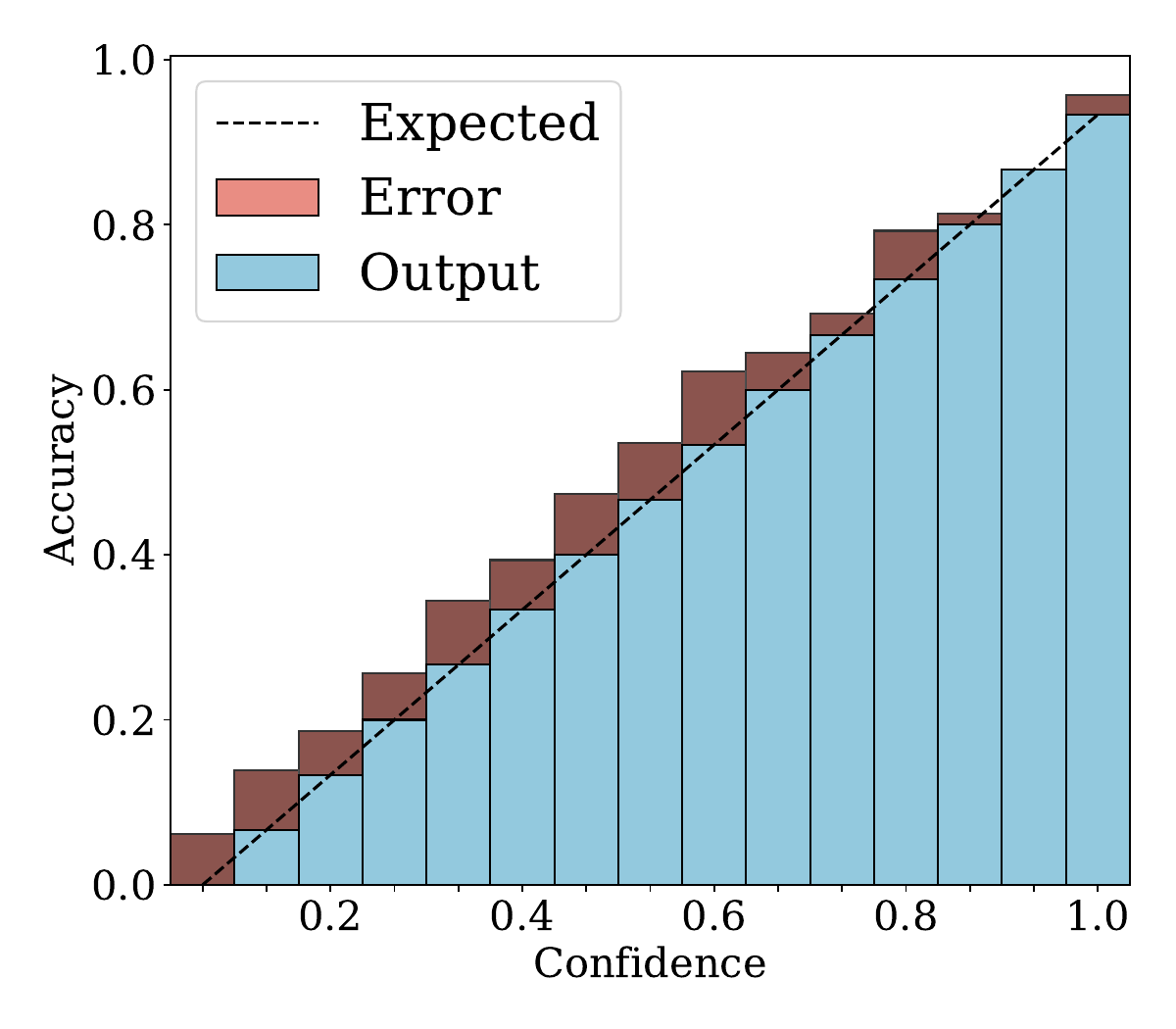} 
       \end{subfigure}  
       \begin{subfigure}{0.291\columnwidth}
           \includegraphics[width=\textwidth]{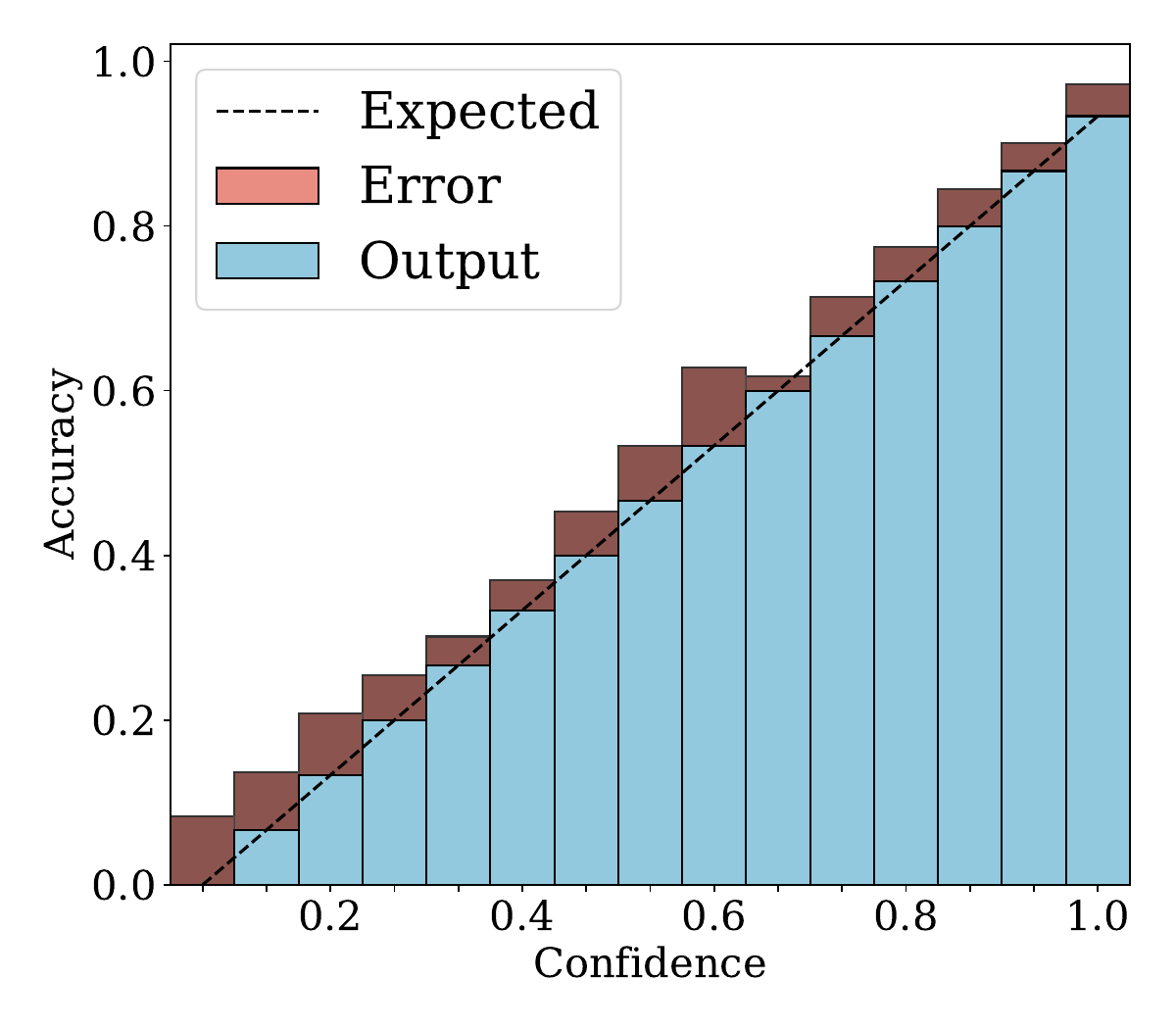}
       \end{subfigure}
       \begin{subfigure}{0.291\columnwidth}
           \includegraphics[width=\textwidth]{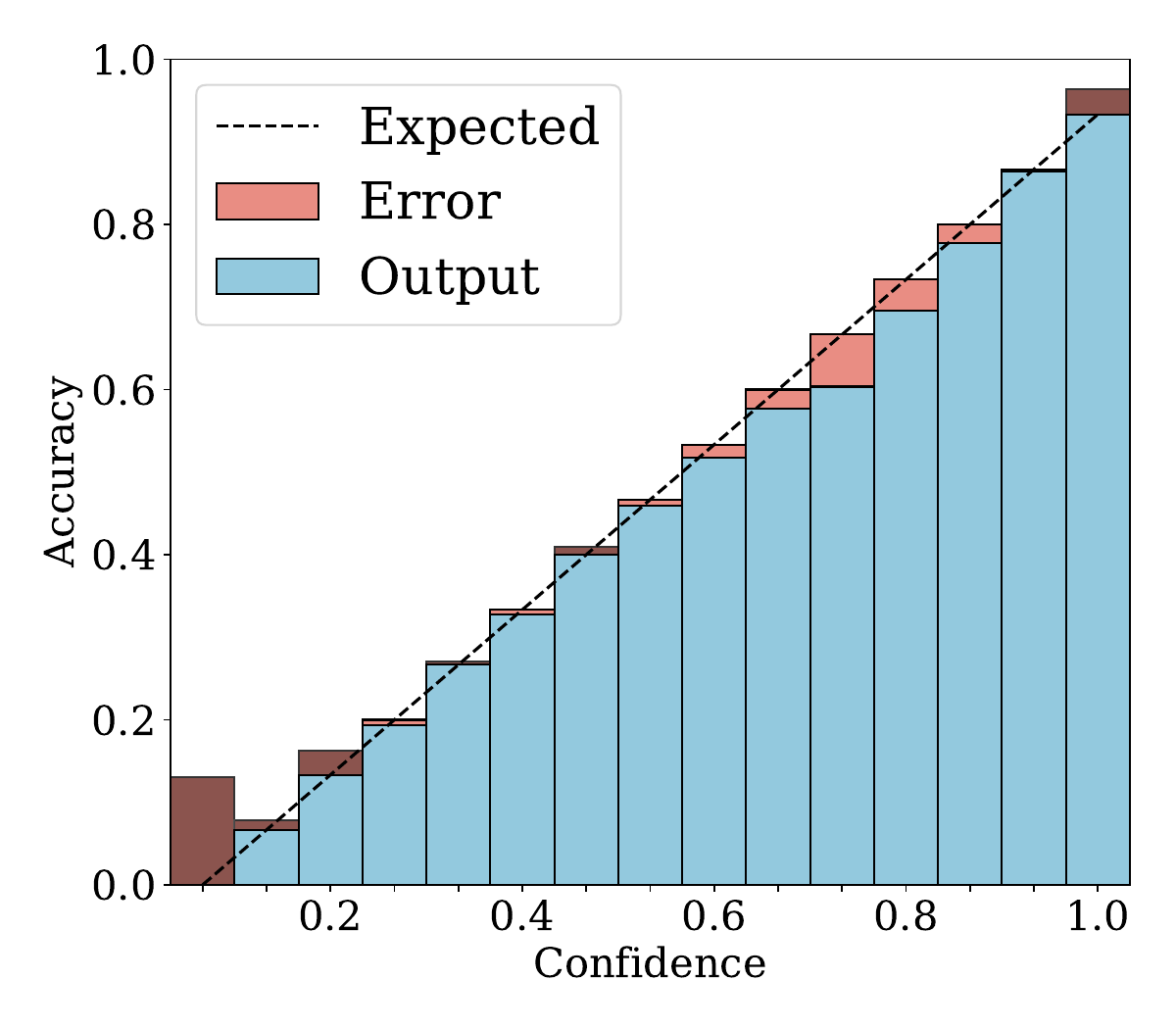}
       \end{subfigure}
       \begin{subfigure}{0.291\columnwidth} 
           \includegraphics[width=\textwidth]{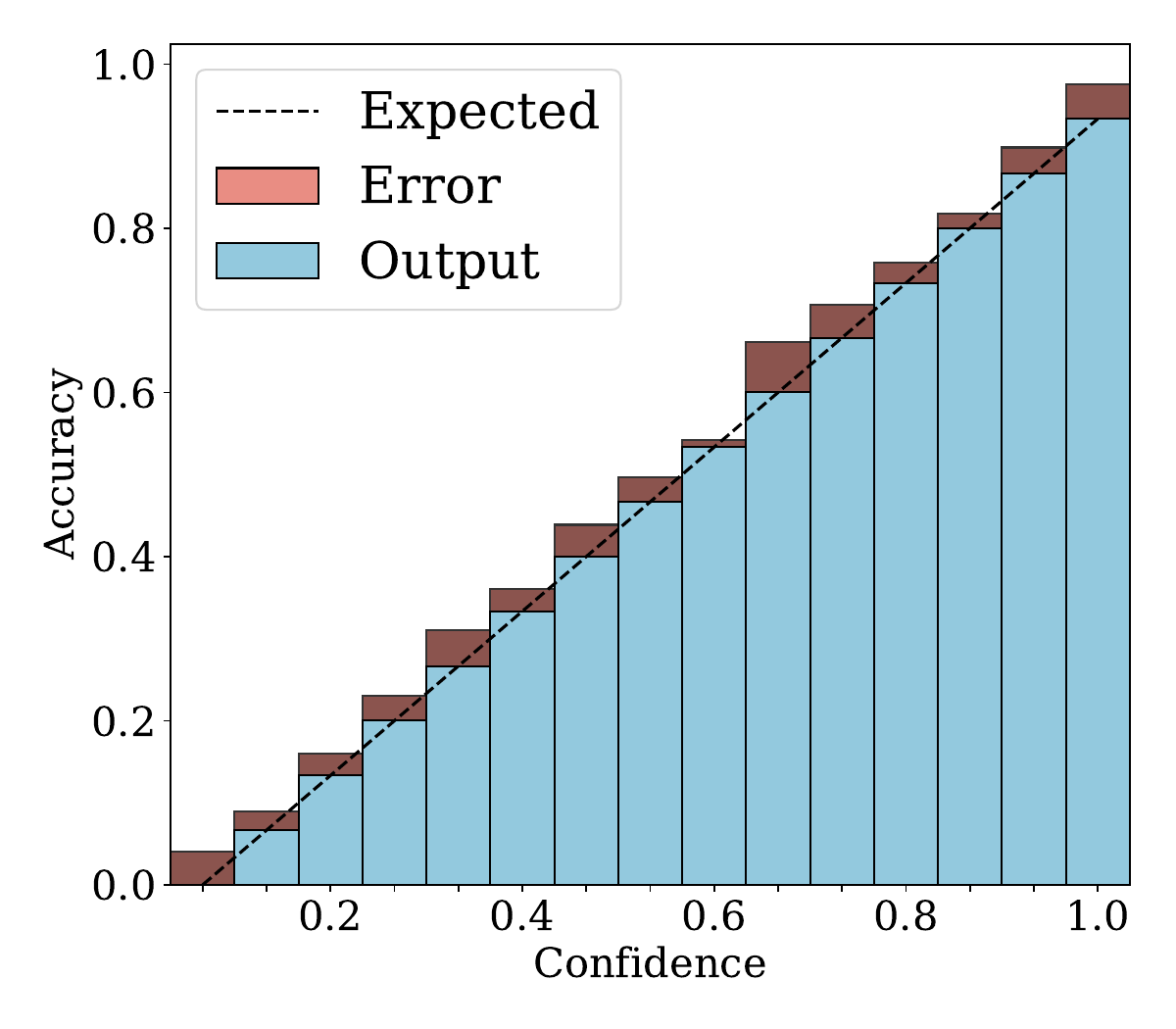} 
       \end{subfigure}  
      
       \vspace{2mm}
       \begin{subfigure}{0.291\columnwidth}
           \includegraphics[width=\textwidth]{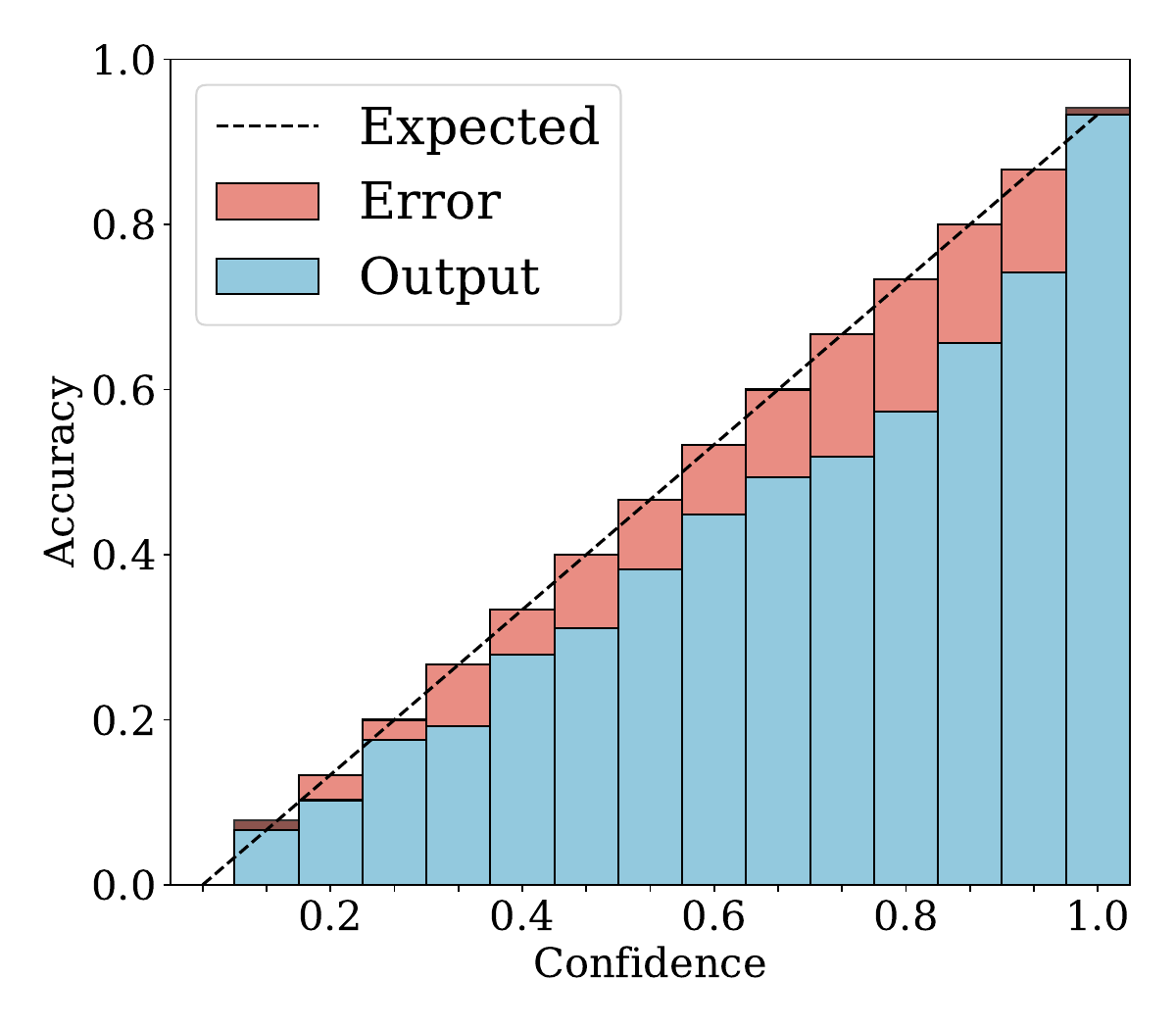}
           \caption{{CE.}}
       \end{subfigure}
       \begin{subfigure}{0.291\columnwidth}
           \includegraphics[width=\textwidth]{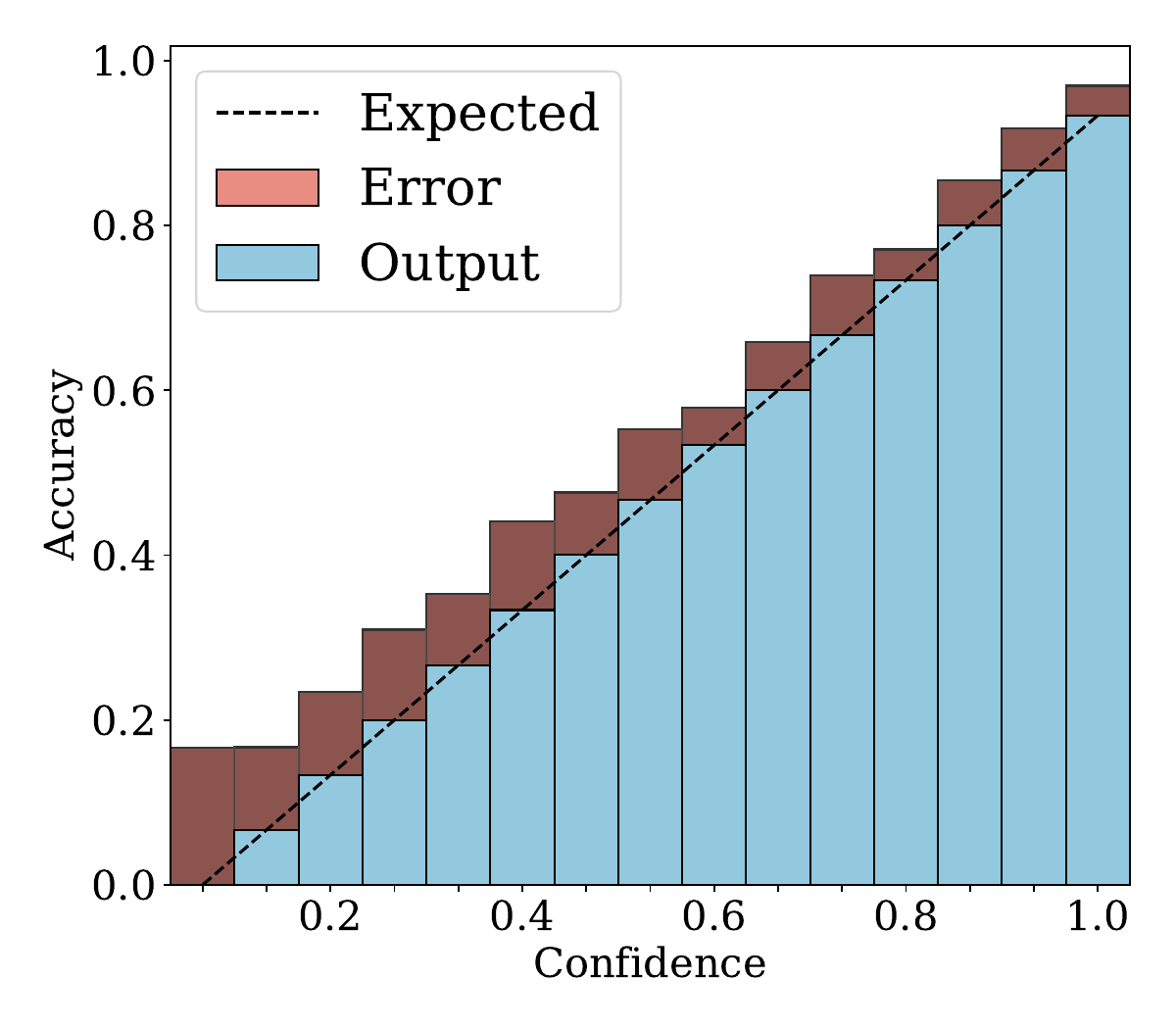}
           \caption{{LS~\cite{szegedy2016rethinking}.}}
       \end{subfigure} 
       \begin{subfigure}{0.291\columnwidth} 
           \includegraphics[width=\textwidth]{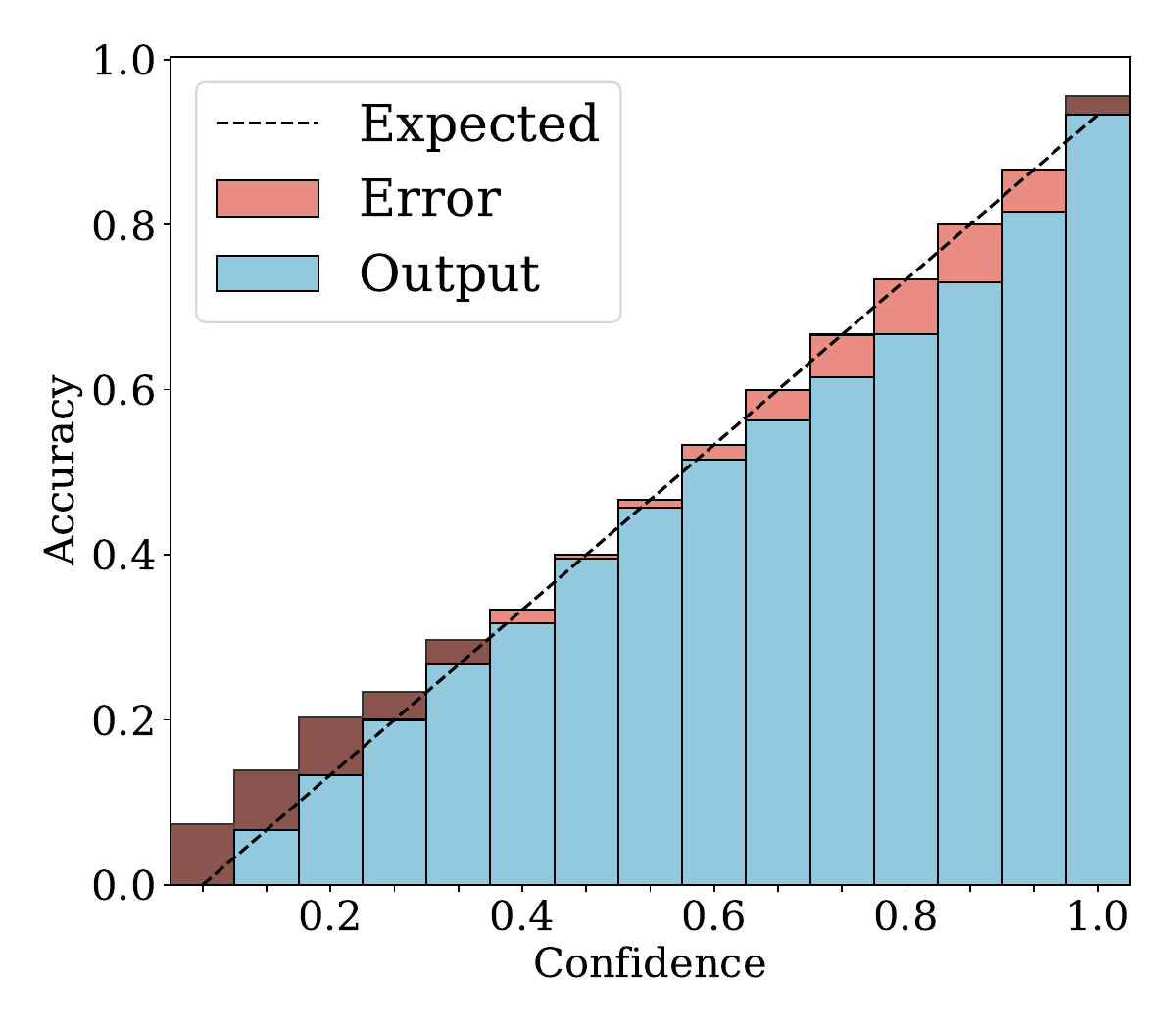} 
           \caption{{CPC~\cite{cheng2022calibrating}.}}
       \end{subfigure}
       \begin{subfigure}{0.291\columnwidth} 
           \includegraphics[width=\textwidth]{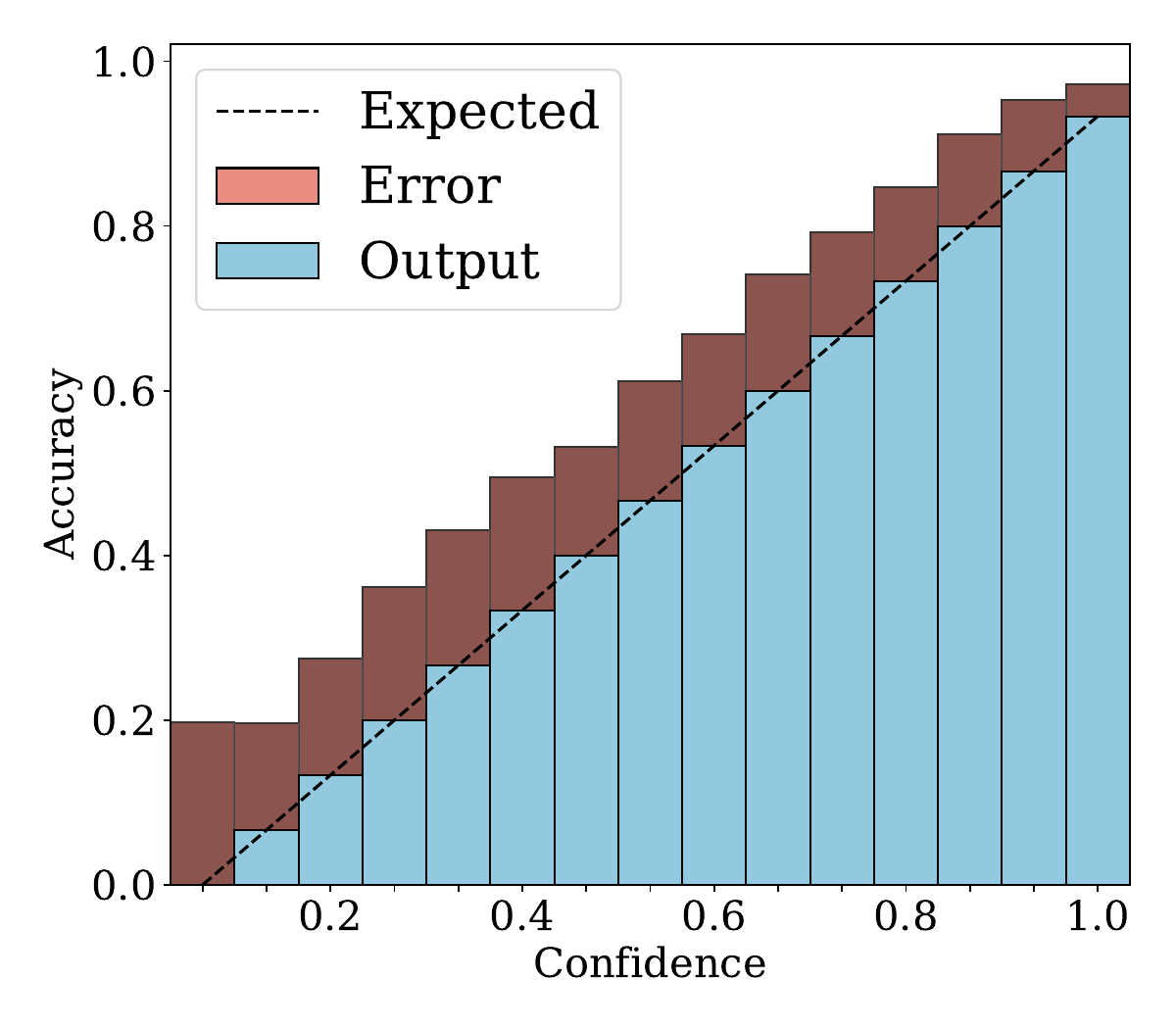}
           \caption{{MDCA~\cite{hebbalaguppe2022stitch}.}}
       \end{subfigure}  
       \begin{subfigure}{0.291\columnwidth}
           \includegraphics[width=\textwidth]{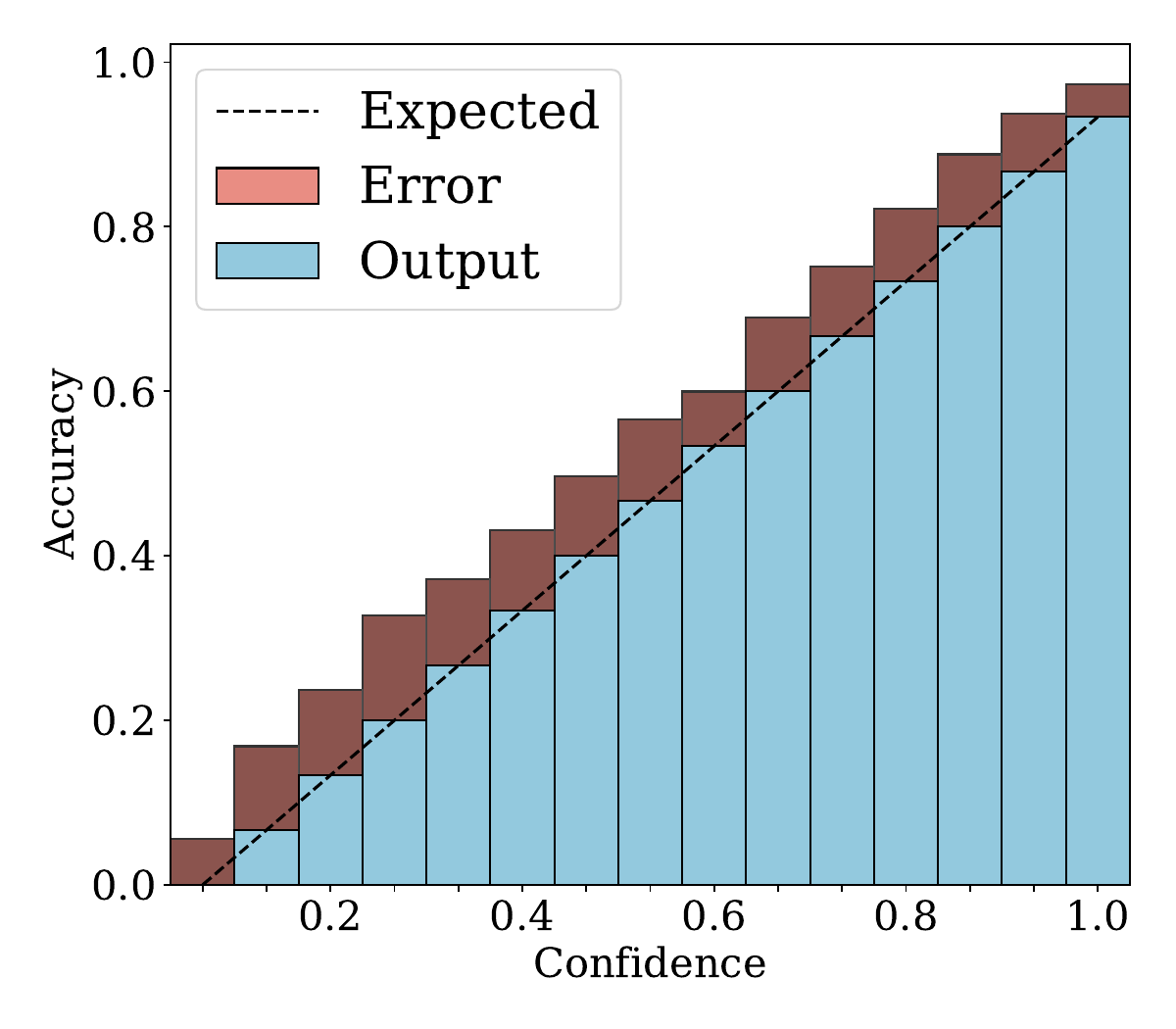}
           \caption{{MbLS~\cite{liu2022devil}.}}
       \end{subfigure}
       \begin{subfigure}{0.291\columnwidth}
           \includegraphics[width=\textwidth]{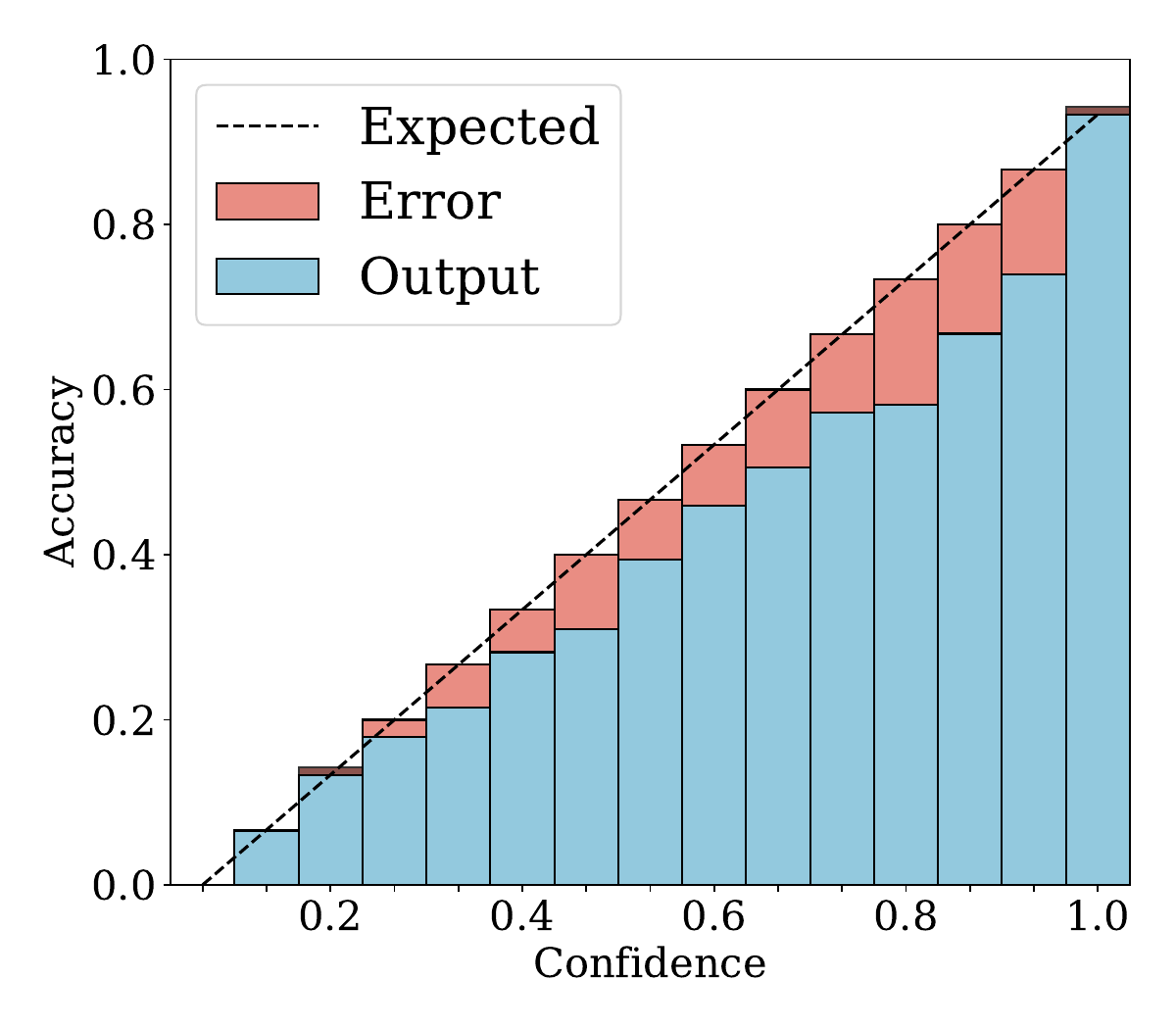}
           \caption{{CRL~\cite{moon2020confidence}.}}
       \end{subfigure} 
       \begin{subfigure}{0.291\columnwidth} 
           \includegraphics[width=\textwidth]{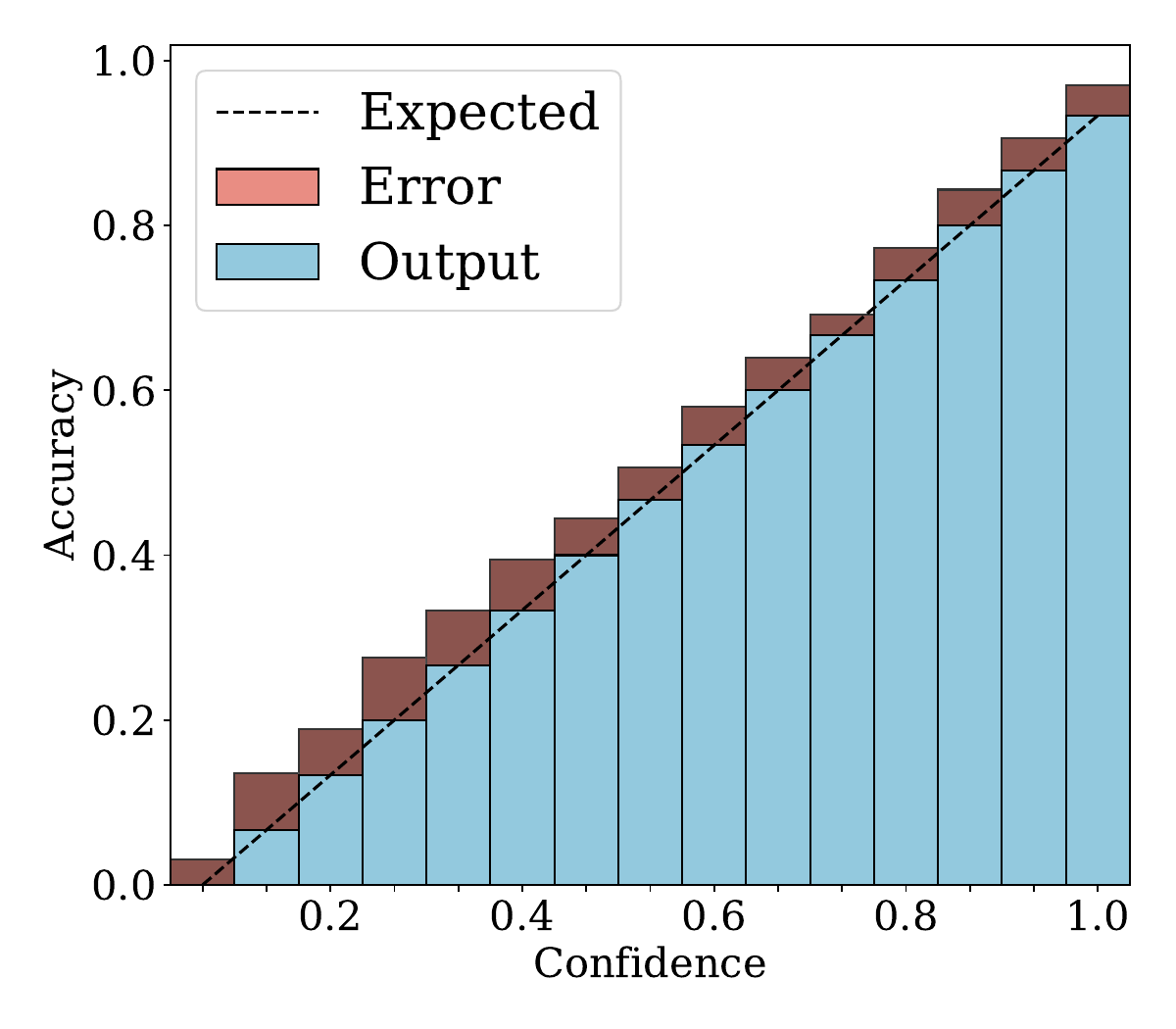}
           \caption{{ACLS.}}
       \end{subfigure}  
   \end{center}
   \vspace{-3mm}
   \captionsetup{font={small}}
   \caption{Comparisons of reliability diagrams on the validation splits of Tiny-ImageNet~\cite{le2015tiny}~(top) and ImageNet~\cite{deng2009imagenet}~(bottom). We exploit ResNet-50~\cite{he2016deep} and compute ECE with 15 bins. We can clearly see that ACLS provides better results than other methods. Best viewed in color.}
   \label{fig:diagram}
   \vspace{-4mm}
\end{figure*}

\subsection{Results} \label{sec:4.2}

\begin{table}[]
   \captionsetup{font={small}}
   \caption{Quantitative results of DeeplabV3~\cite{chen2017rethinking} on the validation set of PASCAL VOC~\cite{everingham2009pascal}. We exploit ResNet-50~\cite{he2016deep} as our backbone, and measure ECE and AECE with 15 bins. Numbers in bold indicate the best performance and underlined ones are the second best.}
   \vspace{-2mm}
   \label{tab:seg}
   \centering
   \small
   \renewcommand{\arraystretch}{1.0}
   \begin{tabular}{l|ccc}
   \hline
   \multicolumn{1}{c|}{Method}        & mIoU $\uparrow$   & ECE $\downarrow$ & AECE $\downarrow$ \\ \hline\hline
   CE                                 & 70.92             & 8.26             & 8.23              \\
   ECP~\cite{pereyra2017regularizing} & \underline{71.16} & 8.31             & 8.26              \\
   LS~\cite{szegedy2016rethinking}    & 71.00             & 9.35             & 9.95              \\
   FL~\cite{lin2017focal}             & 69.99             & 11.44            & 11.43             \\
   MbLS~\cite{liu2022devil}           & \bf{71.20}        & \underline{7.94} & \underline{7.99}  \\ \hline
   ACLS                               & 70.63             & \bf{7.29}        & \bf{7.28}         \\ \hline
   \end{tabular}
   \vspace{-4mm}
\end{table}

We show in Table~\ref{tab:cls} quantitative comparisons between our method with state-of-the-art network calibration methods~\cite{kumar2018trainable,pereyra2017regularizing,szegedy2016rethinking,lin2017focal,mukhoti2020calibrating,moon2020confidence,hebbalaguppe2022stitch,cheng2022calibrating,liu2022devil} on image classification. For CIFAR10~\cite{krizhevsky2009learning} and Tiny-ImageNet~\cite{le2015tiny}, the numbers for the methods~\cite{pereyra2017regularizing,szegedy2016rethinking,lin2017focal,mukhoti2020calibrating,liu2022devil} are taken from MbLS~\cite{liu2022devil}. For a fair comparison, we reproduce the results of MMCE~\cite{kumar2018trainable}, CRL~\cite{moon2020confidence}, CPC~\cite{cheng2022calibrating}, and MDCA~\cite{hebbalaguppe2022stitch} with the same experimental configuration, including network architectures and datasets. For ImageNet~\cite{deng2009imagenet}, we reproduce all the methods in Table~\ref{tab:cls} with ResNet-50~\cite{he2016deep}. From Table~\ref{tab:cls}, we can clearly see that ACLS outperforms all previous calibration methods by significant margins on all benchmarks in terms of ECE and AECE. In particular, we have three findings as follows: (1) ACLS outperforms the AR methods~\cite{hebbalaguppe2022stitch,cheng2022calibrating} by large margins. MDCA~\cite{hebbalaguppe2022stitch} and CPC~\cite{cheng2022calibrating} penalize target labels using adaptive smoothing functions, but they often behave undesirably in terms of network calibration. ACLS addresses this limitation, providing better ECE and AECE, compared with the AR methods. (2) The CR method, MbLS~\cite{liu2022devil}, regularizes confidence values selectively. However, it does not penalize the target labels of predicted classes, outperformed by ACLS on all benchmarks. (3) ACLS alleviates the limitations of AR and CR, and it also surpasses CRL~\cite{moon2020confidence} in terms of ECE and AECE. We provide in Table~\ref{tab:seg} quantitative results of semantic segmentation on PASCAL VOC~\cite{everingham2009pascal}. We can see that our approach outperforms other methods in terms of network calibration, suggesting that it also works well for the dense prediction task.

\subsection{Discussion} \label{sec:4.3}

\paragraph{Reliability diagrams.}
We show in Fig.~\ref{fig:diagram} reliability diagrams on Tiny-ImageNet~\cite{le2015tiny} and ImageNet~\cite{deng2009imagenet}. Following the works of~\cite{mukhoti2020calibrating,cheng2022calibrating,liu2022devil,hebbalaguppe2022stitch,moon2020confidence,guo2017calibration,ghosh2022adafocal,ding2021local,ma2021meta}, we visualize the reliability diagrams of average accuracies and confidences for 15 bins. We can see from Fig.~\ref{fig:diagram} that ACLS outperforms the other methods~\cite{szegedy2016rethinking,liu2022devil,cheng2022calibrating,hebbalaguppe2022stitch,moon2020confidence} across the ranges of confidence. Specifically, we can see from the third, fourth, and last columns, ACLS shows better calibration capability than the AR method~\cite{hebbalaguppe2022stitch,cheng2022calibrating}, since they often behave in opposition to the desirable smoothing function, in terms of network calibration situation~(See Sec.~\ref{sec:3.2}). The fifth and last columns show that the CR method~(MbLS~\cite{liu2022devil}) provides larger values of confidences, compared to ACLS, across all bins. A plausible reason is that it does not penalize target labels when $j = \hat{y}$ as in Eq.~\eqref{eq:ex_mbls}. We can see from the last two columns ACLS surpasses the ACR method~\cite{moon2020confidence} since ACLS alleivates the limitations of AR and CR.

\begin{table}[]
   \captionsetup{font={small}}
   \caption{Comparison of different regularization methods in terms of ECE and AECE on the validation set of Tiny-ImageNet~\cite{le2015tiny}. ECE and AECE are computed with 15 bins. Numbers in bold are the best performance and underlined ones are the second best. We denote by $\bigtriangleup$ adaptive or conditional regularizers with negative effects.}
   \vspace{-1mm}
   \centering
   \small
   \renewcommand{\arraystretch}{1.0}
   \setlength{\tabcolsep}{4pt} 
   \label{tab:reg}
   \begin{tabular}{l|cc|C{0.75cm}C{0.8cm}|C{0.75cm}C{0.8cm}}
   \hline
   \multicolumn{1}{c|}{\multirow{2}{*}{Method}} &
     \multicolumn{1}{c}{\multirow{2}{*}{AR}}    &
     \multicolumn{1}{c|}{\multirow{2}{*}{CR}}   &
     \multicolumn{2}{c|}{ResNet-50}             &
     \multicolumn{2}{c}{ResNet-101}             \\ \cline{4-7} 
   \multicolumn{1}{c|}{}   &
     \multicolumn{1}{c}{}  &
     \multicolumn{1}{c|}{} &
     \scriptsize{~ECE$\downarrow$} &
     \scriptsize{AECE$\downarrow$} &
     \scriptsize{~ECE$\downarrow$} &
     \scriptsize{AECE$\downarrow$} \\ \hline \hline
     LS~\cite{szegedy2016rethinking}    &                  &                  & 3.17             & 3.16             & 2.20             & 2.21               \\ \hline
     MDCA~\cite{hebbalaguppe2022stitch} & $\bigtriangleup$ &                  & 2.77             & 2.61             & 6.06             & 6.06               \\
     Ours~(AR only)                     & $\checkmark$     &                  & 1.54             & \underline{1.42} & \underline{1.34} & \bf{0.89}          \\ \hline
     MbLS~\cite{liu2022devil}           &                  & $\bigtriangleup$ & 1.64             & 1.73             & 1.62             & 1.68               \\ 
     Ours~(CR only)                     &                  & $\checkmark$     & \underline{1.32} & 1.43             & 1.38             & 1.41               \\ \hline
     Ours~(ACLS)                        & $\checkmark$     & $\checkmark$     & \bf{1.05}        & \bf{1.03}        & \bf{1.11}        & \underline{1.15}   \\ \hline
   \end{tabular}
   \vspace{-4mm}
\end{table}

\vspace{-4mm}
\paragraph{Ablation study on AR and CR.}
We show in Table~\ref{tab:reg} an ablation analysis of variant of regularization methods on Tiny-ImageNet~\cite{le2015tiny}. The third and fifth rows show the results of our method that use either a smoothing function of Eq.~\eqref{eq:ours_smoothing} or an indicator function of Eq.~\eqref{eq:ours_decision} only. From the table, we have the following observations: (1) AR and CR outperform LS by significant margins for both networks. This demonstrates that our smoothing and indicator functions in Eqs.~\eqref{eq:ours_smoothing} and~\eqref{eq:ours_decision} alleviate the miscalibration remarkably~(See the first, third, and fifth rows). (2) We can clearly see that our smoothing function alleviates the limitation of previous AR methods~\cite{hebbalaguppe2022stitch,cheng2022calibrating} by large margins~(See the second and third rows). (3) The indicator function in Eq.~\eqref{eq:ours_decision} surpasses LS and MbLS for both architectures. This suggests that our indicator function calibrates the network more effectively than LS and MbLS~(See the first, fourth, and fifth rows). (4) ACLS achieves the best performance, demonstrating that AR and CR are complementary to each other.

\begin{figure}[]
   \begin{center}
       \begin{subfigure}{0.495\columnwidth}
           \includegraphics[width=\textwidth]{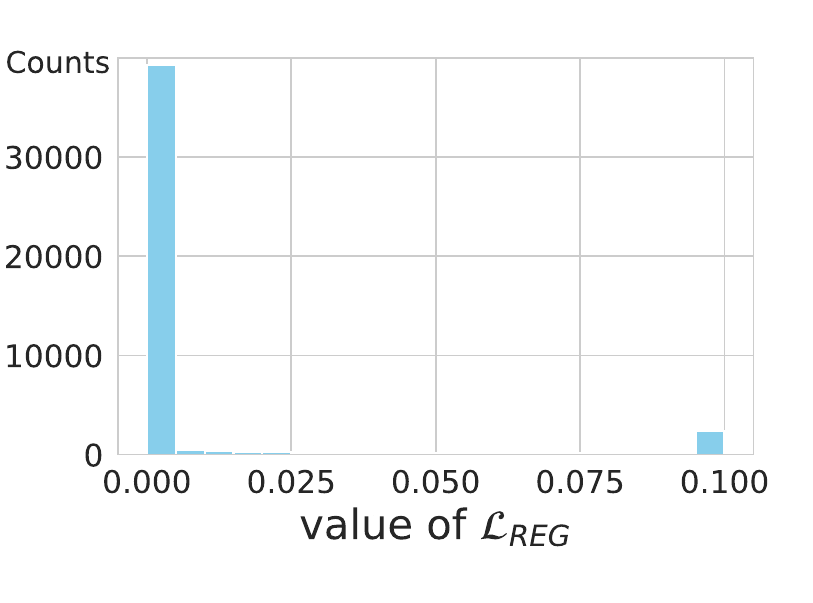}
       \end{subfigure}
       \begin{subfigure}{0.495\columnwidth}
           \includegraphics[width=\textwidth]{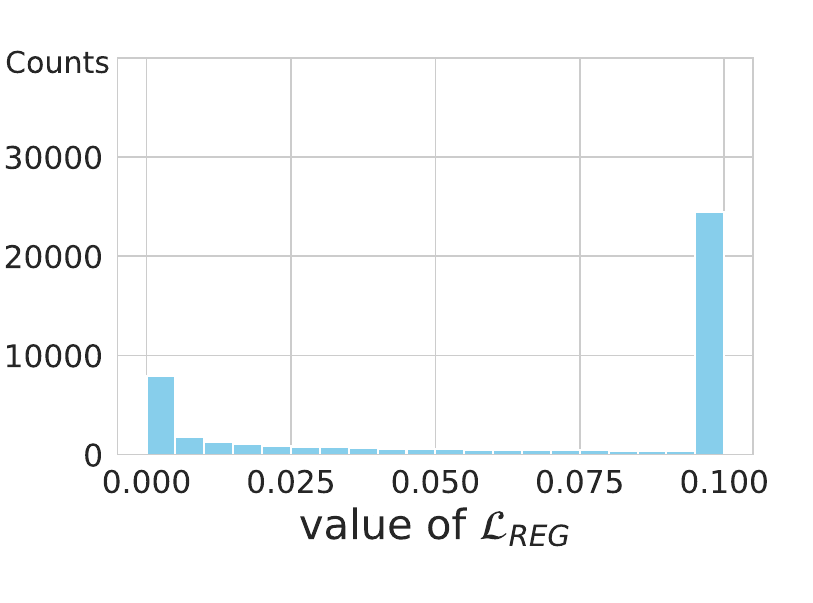}
       \end{subfigure} 
   \end{center}
   \vspace{-4mm}
   \captionsetup{font={small}}
   \caption{Comparisons of $\mathcal{L}_{\text{REG}}$ for the ranking condition~(ACLS with ranking condition, left) and the margin~(ACLS, right). We train ResNet-50~\cite{he2016deep} on Tiny-ImageNet~\cite{le2015tiny}. For visualization, we clip the values of $\mathcal{L}_{\text{REG}}$ into the range $[0.0, 0.1]$ and plot histograms using 20 bins.}
   \label{fig:ranking}
   \vspace{-5mm}
\end{figure}

\vspace{-4mm}
\paragraph{Ablation study on conditions.}
We compare in Table~\ref{tab:condition} ECE and AECE for indicator functions with margin or ranking conditions. To simulate the result in the second row, we set an indicator function as in Eq.~\eqref{eq:ours_decision}, instead of using the original ranking counterpart in CRL~\cite{moon2020confidence}. Similarly, we use the ranking condition in CRL~\cite{moon2020confidence} for the third row. From the table, we have two findings: (1) For both CRL and ACLS, the methods exploiting the margin condition outperform the ranking counterparts consistently, demonstrating that the margin is more effective than the ranking condition in terms of network calibration as described in Sec.~\ref{sec:3.2}. The ranking condition compares the ordinal ranking relationship based on the correctness of each sample, where the correctness values are computed by counting the number of correct classifications for each sample during training. At the late phase of training, each sample has similar correctness value, suggesting that the the ordinal ranking relationship will not be satisfied, and preventing regularization. We can see from Fig.~\ref{fig:ranking} that the regularizer is inactive~(\ie,~$\mathcal{L}_{\text{REG}}=0$) for 85\% of samples, when exploiting ranking conditions~(left), while only for 5\% of samples for the margin condition~(right). Furthermore, at the early phase of training, the correctness history is empty, and thus the networks are trained using softmax CE loss only. This also degrades the calibration of networks. (2) With the same condition, ACLS always outperforms CRL. This demonstrates once again that CRL still suffers from the limitation of AR, while ACLS mitigates the problem. 

\begin{table}[]
   \captionsetup{font={small}}
   \caption{Quantitative comparison of indicator functions exploiting margin or ranking conditions on the validation split of Tiny-ImageNet~\cite{le2015tiny}. We use ResNet-50~\cite{he2016deep} and compute ECE and AECE using 15 bins. Numbers in bold indicate the best performance and underlined ones are the second best.}
   \vspace{-1mm}
   \centering
   \label{tab:condition}
   \renewcommand{\arraystretch}{1.0}
   \setlength{\tabcolsep}{4pt} 
   \small
   \begin{tabular}{l|cc|cc}
   \hline
                                                   & Ranking      & Margin       & ~ECE$\downarrow$ & AECE$\downarrow$ \\ \hline\hline
   \multirow{2}{*}{CRL~\cite{moon2020confidence}}  & $\checkmark$ &              & 1.65             & 1.52             \\
                                                   &              & $\checkmark$ & \underline{1.47} & 1.40             \\ \hline
   \multirow{2}{*}{Ours~(ACLS)}                    & $\checkmark$ &              & 1.48             & \underline{1.19} \\
                                                   &              & $\checkmark$ & \bf{1.05}        & \bf{1.03}        \\ \hline
   \end{tabular}
   \vspace{-3mm}
\end{table}

\vspace{-4mm}
\paragraph{Limitation.} Similar to other regularization-based calibration methods~\cite{cheng2022calibrating,hebbalaguppe2022stitch,lin2017focal,liu2022devil,mukhoti2020calibrating,moon2020confidence}, ACLS requires re-training networks from scratch, which is computationally expensive. Also, ACLS might not behave as expected for an exceptional case. Let us suppose the case of $z_j > z_k > z_i$,~\ie,~$\min \mathbf{z}=z_i$ and $\max \mathbf{z}=z_j$, and assume that $z_j - z_i$ and $z_j - z_k$ are sufficiently large and small, respectively, and $y=j$. ACLS tries to reduce $z_j$, since $z_j - z_i$ is large enough. The ordering for~$z_j$ and $z_k$ could then be altered,~\ie,~$z_j < z_k$, which in turn changes the prediction after a backward pass~(\ie, $\hat{y}=k$ from $j$). This might degrade the classification performance. We have observed that this altering occurs in 0.29\% of samples during the last epoch of training on Tiny-ImageNet~\cite{le2015tiny}.

\section{Conclusion}
We have provided an in-depth analysis on existing calibration methods, and have found that they can be interpreted as variants of label smoothing. Based on this, we have presented a new regularization term for network calibration, dubbed ACLS, that addresses both overconfidence and underconfidence problems effectively. Finally, we have shown that our approach sets a new state of the art on standard calibration benchmarks.

\vspace{2mm}
\noindent \small{\textbf{Acknowledgments.} This work was partly supported by IITP grant funded by the Korea government (MSIT) (No.RS-2022-00143524, Development of Fundamental Technology and Integrated Solution for Next-Generation Automatic Artificial Intelligence System, No.2022-0-00124, Development of Artificial Intelligence Technology for Self-Improving Competency-Aware Learning Capabilities) and the KIST Institutional Program (Project No.2E31051-21-203).}

{\small
\bibliographystyle{ieee_fullname}
\bibliography{egbib}
}

\clearpage


\section*{Supplementary material}

# ACLS: Adaptive and Conditional Label Smoothing for Network Calibration Supplement

Hyekang Park[1]    Jongyoun Noh[1]    Youngmin Oh[1]
Donghyeon Baek[1]    Bumsub Ham[1,2*]

[1]Yonsei University    [2]Korea Institute of Science and Technology (KIST)

https://cvlab.yonsei.ac.kr/projects/ACLS

In this supplementary material, we first present detailed derivations of gradients for previous regularization-based calibration methods (Sec. S1). We then describe more details for calibration metrics and hyperparameters (Sec. S2). We also provide more results on our approach (Sec.S3).

## S1. Gradient analysis

### S1.1. FLSD

MbLS [14] has shown that a focal loss (FL) [13] can be regarded as a form of label smoothing (LS) [21]. Taking one step further, we observe that FLSD [18] can also be approximated as LS. Specifically, FLSD uses a sample-dependent focusing parameter $\gamma$ to improve calibration of FL. $\gamma$ is determined by a heuristic rule as follows:

$$\gamma = \begin{cases} 5, & p_{\hat{y}} \in [0, 0.2) \\ 3, & p_{\hat{y}} \in [0.2, 1) \end{cases}. \quad (1)$$

We empirically find that $\gamma = 3$ for 95% of samples on Tiny-ImageNet [12]. This suggests that we can approximate FLSD [18] as FL and thus it can be viewed in the form of LS. Based on the approximation in [14] and our observation, we formulate smoothing and indicator functions of FLSD as in Table 2 in the main paper.

### S1.2. CPC

**Gradients.** CPC [1] calibrates $C(C-1)/2$ binary pairs of probabilities instead of regularizing original $C$-way softmax probabilities. The loss function of CPC is as follows:

$$\mathcal{L} = \mathcal{L}_{\text{CE}} + \mathcal{L}_{\text{CPC}} = \mathcal{L}_{\text{CE}} + \lambda_1 \mathcal{L}_{\text{1v1}} + \lambda_2 \mathcal{L}_{\text{BE}} \quad (2)$$

where $\lambda_1$ and $\lambda_2$ are hyperparameters. $\mathcal{L}_{\text{1v1}}$ and $\mathcal{L}_{\text{BE}}$ are defined as follows:

$$\mathcal{L}_{\text{1v1}} = -\sum_{k \neq y}^{C} \log \frac{p_y}{p_y + p_k}, \quad (3)$$

and

$$\mathcal{L}_{\text{BE}} = -\sum_{k,l \neq y}^{C} \left( \log \frac{p_k}{p_k + p_l} + \log \frac{p_l}{p_k + p_l} \right). \quad (4)$$

We compute the gradients of $\mathcal{L}_{\text{1v1}}$ and $\mathcal{L}_{\text{BE}}$ w.r.t $z_j$ as follows:

$$\begin{aligned}
\frac{\partial \mathcal{L}_{\text{1v1}}}{\partial z_j} &= \frac{\partial}{\partial z_j} \left( -\sum_{k \neq y}^{C} \log \frac{p_y}{p_y + p_k} \right) \\
&= -\sum_{k \neq y} \frac{p_y + p_k}{p_y} \frac{\frac{\partial p_y}{\partial z_j}(p_y + p_k) - p_y\left(\frac{\partial p_y}{\partial z_j} + \frac{\partial p_k}{\partial z_j}\right)}{(p_y + p_k)^2} \\
&= \begin{cases} -\sum_{k \neq y} \frac{p_k}{p_j + p_k}, & j = y \\ \frac{p_j}{p_y + p_j}, & j \neq y \end{cases},
\end{aligned} \quad (5)$$

and

$$\begin{aligned}
\frac{\partial \mathcal{L}_{\text{BE}}}{\partial z_j} &= \frac{\partial}{\partial z_j} \left( -\sum_{k,l \neq y}^{C} \left( \log \frac{p_k}{p_k + p_l} + \log \frac{p_l}{p_k + p_l} \right) \right) \\
&= -\sum_{k,l \neq y} \frac{p_l + p_k}{p_k} \frac{\frac{\partial p_k}{\partial z_j}(p_k + p_l) - p_k\left(\frac{\partial p_k}{\partial z_j} + \frac{\partial p_l}{\partial z_j}\right)}{(p_k + p_l)^2} \\
&\quad - \sum_{k,l \neq y} \frac{p_k + p_l}{p_l} \frac{\frac{\partial p_l}{\partial z_j}(p_l + p_k) - p_l\left(\frac{\partial p_l}{\partial z_j} + \frac{\partial p_k}{\partial z_j}\right)}{(p_l + p_k)^2} \\
&= -2 \sum_{k,l \neq y} \frac{\frac{\partial p_k}{\partial z_j}(p_k + p_l) - p_k\left(\frac{\partial p_k}{\partial z_j} + \frac{\partial p_l}{\partial z_j}\right)}{(p_k + p_l)p_k} \\
&= \begin{cases} 0, & j = y \\ 2\sum_{k \neq y} \frac{p_j - p_k}{p_j + p_k}, & j \neq y \end{cases}.
\end{aligned} \quad (6)$$

---

[*]Corresponding author

We compute the gradients of CPC w.r.t $z_j$ using Eqs. (5) and (6) as follows:

$$\frac{\partial \mathcal{L}}{\partial z_j} = \frac{\partial \mathcal{L}_{\text{CE}}}{\partial z_j} + \frac{\partial \mathcal{L}_{\text{CPC}}}{\partial z_j} = \frac{\partial \mathcal{L}_{\text{CE}}}{\partial z_j} + \lambda_1 \frac{\partial \mathcal{L}_{\text{1v1}}}{\partial z_j} + \lambda_2 \frac{\partial \mathcal{L}_{\text{BE}}}{\partial z_j}$$

$$= \begin{cases} p_j - \left(q_j + \lambda_1 \sum_{k \neq j} \frac{p_k}{p_j + p_k}\right) & j = y \\ p_j - \left(q_j + \lambda_1 \frac{p_j}{p_y + p_j} + \lambda_2 \sum_{k \neq y} \frac{p_k - p_j}{p_k + p_j}\right) & j \neq y \end{cases}. \tag{7}$$

where $\lambda_1$ and $\lambda_2$ are hyperparameters. We omit the constant (*e.g.*, 2) for brevity.

**Smoothing and indicator functions.** Reformulating Eq. (7), we define a smoothing function of CPC as follows:

$$f(z_j) = \begin{cases} -\lambda_1 \sum_{k \neq j}^{C} \frac{p_k}{p_k + p_j}, & j = y \\ \lambda_1 \frac{p_j}{p_y + p_j} + \lambda_2 \sum_{k \neq y} \frac{p_j - p_k}{p_j + p_k}, & j \neq y \end{cases}. \tag{8}$$

Eq. (8) shows that CPC adjusts the labels and the only difference between CPC and LS [21] is how they determine the degree of smoothing. For example, LS smoothes using fixed constants, while CPC determines the degree of smoothing based on the probabilities. Note that an indicator function of CPC is always 1.

## S1.3. MDCA

**Gradients.** MDCA [9] presents a differentiable version of ECE that approximates ECE in a mini-batch. MDCA exploits it for a regularizer directly. The loss function is defined as follows:

$$\mathcal{L} = \mathcal{L}_{\text{CE}} + \mathcal{L}_{\text{MDCA}}$$
$$= \mathcal{L}_{\text{CE}} + \lambda \sum_{k=1}^{C} \frac{1}{N} \left| \sum_{n=1}^{N} p_i^{(n)} - \sum_{n=1}^{N} q_i^{(n)} \right|, \tag{9}$$

where $\lambda$ is a hyperparameter and $n$ is an index of an input sample. $\frac{1}{N} \sum_{n=1}^{N} q_i^{(n)}$ represents a ratio of input samples for class $i$ in a mini-batch. Thus, $\mathcal{L}_{\text{MDCA}}$ in Eq. (9) encourages $p_i$ to have the same value of the ratio for class $i$. We compute the gradients of Eq. (9) w.r.t $z_j$ as follows:

$$\frac{\partial \mathcal{L}}{\partial z_j^{(n)}} = \frac{\partial \mathcal{L}_{\text{CE}}}{\partial z_j^{(n)}} + \frac{\partial \mathcal{L}_{\text{MDCA}}}{\partial z_j^{(n)}}$$
$$= p_j^{(n)} - q_j^{(n)} + \frac{\lambda}{N} \sum_k a_k \frac{\partial p_k^{(n)}}{\partial z_j^{(n)}}$$
$$= p_j^{(n)} - q_j^{(n)} + \frac{\lambda}{N} a_j p_j^{(n)} - \frac{\lambda}{N} \sum_{k=1}^{C} a_k p_k^{(n)} p_j^{(n)}$$
$$= p_j^{(n)} - \left( q_j^{(n)} - \frac{\lambda}{N} p_j^{(n)} \left( a_j - \sum_{k=1}^{C} a_k p_k^{(n)} \right) \right). \tag{10}$$

We define $a_k = \text{sgn}\left(\frac{1}{N} \left| \sum_{n=1}^{N} p_k^{(n)} - \sum_{n=1}^{N} q_k^{(n)} \right|\right)$, where we denote by $\text{sgn}(\cdot)$ a sign function whose value is 1 if the argument is positive and $-1$ otherwise. If $p_{\hat{y}}$ is overconfident, while $p_j$ ($j \neq \hat{y}$) are underconfident (*i.e.*, $a_j$ is 1 if $j = \hat{y}$ and $-1$ otherwise.), Eq. (10) reduces as follows:

$$\frac{\partial \mathcal{L}}{\partial z_j^{(n)}} = \begin{cases} p_j - (q_j - \lambda p_j (1 - \sum_k a_k p_k)), & j = \hat{y} \\ p_j - (q_j + \lambda p_j (1 + \sum_k a_k p_k)), & j \neq \hat{y} \end{cases}, \tag{11}$$

where we omit $n$ and $N$ for brevity.

**Smoothing and indicator functions.** We define a smoothing function of MDCA from Eq. (11) as follows:

$$f(z_j) = \begin{cases} \lambda p_j (1 - \sum_k a_k p_k), & j = \hat{y} \\ \lambda p_j (1 + \sum_k a_k p_k), & j \neq \hat{y} \end{cases}. \tag{12}$$

We observe that $|\sum_{k \neq \hat{y}} a_k p_k| \ll |a_{\hat{y}} p_{\hat{y}}|$ and thus we can formulate Eq. (12) as the fourth row in Table 2 in the main paper as follows:

$$f(z_j) = \begin{cases} \lambda p_j (1 - p_j), & j = \hat{y} \\ \lambda p_j (1 + p_{\hat{y}}), & j \neq \hat{y} \end{cases}. \tag{13}$$

Eq. (13) suggests that MDCA reduces the labels if the probability for $j = \hat{y}$ is overconfident, and raises the labels if the probabilities for $j \neq \hat{y}$ are underconfident. For cases other than specified in Eq. (11) (*i.e.*, $a_j$ is 1 if $j = \hat{y}$ and $-1$ otherwise.), Eq. (10) still reduces the label if $p_j$ is higher than the ratio for class $j$, and vice versa if $p_j$ is lower than the ratio. Note that an indicator function of MDCA is always 1.

## S1.4. MbLS

**Gradients.** While LS [21] adjusts $q_j$ for all $j$, MbLS [14] regularizes each label selectively by exploiting a margin $M$. Specifically, its loss function is as follows:

$$\mathcal{L} = \mathcal{L}_{\text{CE}} + \mathcal{L}_{\text{MbLS}}$$
$$= \mathcal{L}_{\text{CE}} + \lambda \sum_{k=1}^{C} \max(0, z_{\hat{y}} - z_k - M), \tag{14}$$

When $z_{\hat{y}} - z_j - M \geq 0$, we compute its gradients w.r.t $z_j$ as follow:

$$\frac{\partial \mathcal{L}}{\partial z_j} = \frac{\partial \mathcal{L}_{\text{CE}}}{\partial z_j} + \frac{\partial \mathcal{L}_{\text{MbLS}}}{\partial z_j}$$
$$= p_j - q_j + \lambda \frac{\partial}{\partial z_j} \left( \sum_{k=1}^{C} z_{\hat{y}} - z_k - M \right) \tag{15}$$
$$= \begin{cases} p_j - q_j, & j = \hat{y} \\ p_j - (q_j + \lambda), & j \neq \hat{y} \end{cases}.$$

**Smoothing and indicator functions.** Note that MbLS penalizes the label using a constant (*e.g.*, $\lambda$). Thus, we can define a smoothing function as follows:

$$f(z_j) = \lambda. \qquad (16)$$

Considering that Eq. (15) reduces to the gradients of CE if $z_{\hat{y}} - z_j - M < 0$, we define an indicator function of MbLS as follows:

$$\mathbb{C}(z_j) = \begin{cases} 0, & j = \hat{y} \\ \mathbb{1}[z_{\hat{y}} - z_j \geq M], & j \neq \hat{y} \end{cases}. \qquad (17)$$

## S1.5. CRL

**Gradients.** CRL [17] formulates network calibration as an ordinal ranking problem [2]. It determines whether to apply regularization by using a ranking condition. Its loss function is as follows:

$$\begin{aligned}\mathcal{L} &= \mathcal{L}_{\text{CE}} + \mathcal{L}_{\text{CRL}} \\ &= \mathcal{L}_{\text{CE}} + \lambda \sum_{n,m} \max\left(0, -\text{sgn}(H(n,m))(p_{\hat{y}}^{(n)} - p_{\hat{y}}^{(m)})\right).\end{aligned}$$
$$(18)$$

where $n$ and $m$ are indices of input samples in the training dataset. We define $H(n,m) = h^{(n)} - h^{(m)}$, where $h^{(n)}$ stores a ranking history of the $n$-th sample in a training dataset. In practice, $n$ and $m$ are selected only in a mini-batch and $\mathcal{L}_{\text{CRL}}$ is calculated using a ranking loss [22]. When the ranking condition is satisfied (*e.g.*, $h^{(n)} < h^{(m)}$ but $p_{\hat{y}}^{(n)} \geq p_{\hat{y}}^{(m)}$), $\mathcal{L}_{\text{CRL}}$ has a non-zero value. In this case, we can compute the gradients of Eq. (18) w.r.t $z_j^{(n)}$ for the specific $n$ as follows:

$$\begin{aligned}\frac{\partial \mathcal{L}}{\partial z_j^{(n)}} &= \frac{\partial \mathcal{L}_{\text{CE}}}{\partial z_j^{(n)}} + \frac{\partial \mathcal{L}_{\text{CRL}}}{\partial z_j^{(n)}} \\ &= p_j^{(n)} - q_j^{(n)} + \lambda \frac{\partial p_{\hat{y}}^{(n)}}{\partial z_j^{(n)}} \\ &= \begin{cases} p_j^{(n)} - \left(q_j^{(n)} - \lambda p_j^{(n)}(1 - p_j^{(n)})\right), & j = \hat{y} \\ p_j^{(n)} - \left(q_j^{(n)} + \lambda p_{\hat{y}}^{(n)} p_j^{(n)}\right), & j \neq \hat{y} \end{cases},\end{aligned}$$
$$(19)$$

since $-\text{sgn}(H(n,m)) = 1$ and $\frac{\partial p_{\hat{y}}^{(m)}}{\partial z_j^{(n)}} = 0$.

**Smoothing and indicator functions.** Based on Eq. (19), we first define a smoothing function of CRL as follows:

$$f(z_j) = \begin{cases} \lambda p_j(1 - p_j), & j = \hat{y} \\ \lambda p_{\hat{y}} p_j, & j \neq \hat{y} \end{cases}, \qquad (20)$$

where we omit $n$ for brevity. Considering that the gradients of Eq. (19) are the same for those of CE if the ranking condition is not satisfied, we can define its indicator function as follows:

$$\mathbb{C}(z_j) = \mathbb{1}\left[H(n,m)(p_{\hat{y}}^{(n)} - p_{\hat{y}}^{(m)}) < 0\right]. \qquad (21)$$

CRL adjusts the labels based on the values of probability (Eq. (20)) only when the condition is satisfied (Eq. (21)). Thus, we can view CRL as a combination of AR and CR (ACR). Combining Eqs (20) and (21), we can obtain the gradients of CRL w.r.t $z_j$ for the specific $n$ in the form of Eq. (8) in the main paper.

## S2. More details

### S2.1. Metrics

Following [1, 4, 6, 7, 9, 14, 16, 17, 18, 19], we report expected calibration error (ECE) and adaptive ECE (AECE) as calibration metrics. ECE is defined as follows:

$$\text{ECE} = \mathbb{E}_{\mathbf{p}}\left[P\left(\hat{y} = y | p_{\hat{y}}\right) - p_{\hat{y}}\right]. \qquad (22)$$

In practice, we approximate Eq. (22) by partitioning predictions into bin $B$. Specifically, we first divide a range $(0, 1]$ into $\mathcal{M}$ equidistance bins (*e.g.*, if $\mathcal{M} = 5$, the bins are $(0, 0.2], (0, 0.4], \cdots, (0.8, 1.0]$). Given a sample, we then assign the sample to the specific bin according to the value of prediction for each sample. For example, if the value of prediction is $0.9$ and $\mathcal{M} = 5$, the sample is assigned to the fifth bin. Following this protocol, we reformulate Eq. (22) as follows:

$$\text{ECE} = \sum_{i=1}^{\mathcal{M}} \frac{|B_i|}{N} \left|\text{acc}\left(B_i\right) - \text{conf}\left(B_i\right)\right|, \qquad (23)$$

where $B_i$ is the $i$-th bin and $N$ is the number of samples. To visualize reliability diagrams, we compute the average accuracies and the calibration errors in each bin. For computing AECE, we partition the range $(0, 1]$ into $\mathcal{M}$ bins, each of which includes the same number of samples, instead of dividing the range into the equidistance bins.

### S2.2. Hyperparameters

To set the hyperparameters $\lambda_1$ and $\lambda_2$ of Eq. (10) in the main paper, we perform a grid search on the cross-validation split of Tiny-ImageNet [12]: $\lambda_1 \in \{0, 0.05, 0.1, 0.15\}$ and $\lambda_2 \in \{0, 0.005, 0.01, 0.015\}$. For ImageNet [3], we use the hyperparameters obtained from Tiny-ImageNet since the search on ImageNet is computationally demanding. We summarize in Table A the results of grid search on the cross-validation split of Tiny-ImageNet. We set $\lambda_1$ to $0.1$ and $\lambda_2$ to $0.01$, respectively. Following [14], we set the margin $M$ to $10$. We also provide the ablation study for $M$ in Sec. S3.

Table A: Comparisons of ACC (%) and ECE (%) on the cross-validation split of Tiny-ImageNet [12] (ACC/ECE). We train ResNet-50 accordring to the values of $\lambda_1$ and $\lambda_2$. ECE is computed with 15 bins. -: Fail to converge.

| $\lambda_1$ \ $\lambda_2$ | 0 | 0.005 | 0.01 | 0.015 |
|---|---|---|---|---|
| 0 | 64.62 / 4.39 | 64.43 / 3.42 | 55.81 / 2.94 | -/- |
| 0.05 | 64.83 / 2.36 | 64.83 / 1.95 | 64.76 / 1.73 | 64.48 / <u>1.41</u> |
| 0.1 | 65.43 / 2.67 | 64.74 / 1.94 | 65.40 / **1.31** | 65.17 / 1.52 |
| 0.15 | 65.04 / 2.71 | 65.18 / 1.44 | 65.23 / 1.42 | 64.11 / 1.47 |

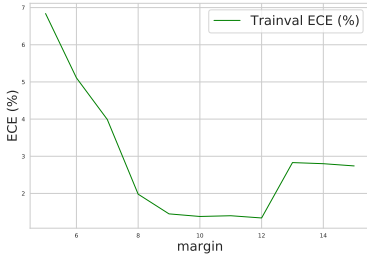

Figure A: Comparisons of ECE accordring to the value of $M$ on the cross-validation (trainval) split of Tiny-ImageNet [12]. We use ResNet-50 [8] and compute ECE with 15 bins.

## S3. More results and discussions

**Effects of margins.** We show in Fig. A ECE accordring to the value of $M$ on the cross-validation split of Tiny-ImageNet [12]. From the figure, we have two findings: (1) If the margin is too small, the network is poorly calibrated. A plausible reason is that the condition is easily satisfied, and thus the regularization term penalizes well-calibrated probabilities also. (2) When the margin is too large, ECE is degraded also. This is because the margin condition is not satisfied even for the miscalibrated probabilities, disturbing calibration.

**Additional results.** We show in Table B quantitative comparisons with our method and other calibration approaches, including temperature scaling (TS) [7], MMCE [11], CutMix [10], CaPE-bin [15], and MCdrop [5], on image classification. For a fair comparison, we have reproduced all results with the same experimental configuration, including a network architecture and a dataset, except for the numbers of Patra et al. [20] which are taken from the paper. From Table B, we can clearly see that ACLS outperforms all methods by a significant margin in terms of ECE and AECE.

**Comparisons with other non-label smoothing-based methods.** We show quantitative comparisons with our method, CutMix [10], CaPE-bin [15], and MCdrop [5] in Table B. CutMix is a data augmentation strategy that replaces a region of an input image with a patch from another one. While CutMix is originally introduced to improve the generalization ability of neural networks, it has proven to be effective for network calibration. However, it often removes salient regions of images and produces inappropriate examples, which might degrade the discriminative ability of networks. CaPE is a regularization-based calibration method that exploits an accuracy of each sample directly. It assigns samples to particular bins and exploits accuracies for the bins (empirical accuracy) as target labels for network calibration, optimizing networks directly in terms of ECE. Computing the empirical accuracies of samples is however computationally demanding. MCdrop approximates the Bayesian inference with multiple predictions using dropout layers. This reduces the computational cost for uncertainty estimation, but still requires several forward passes for predictions. On the contrary, our method provides well-calibrated predictions in a single forward pass.

Table B: Quantitative results on the validation split of Tiny-ImageNet [12] in terms of ACC, ECE, and AECE. We compute the calibration metrics with both 10 and 15 bins using ResNet-50 [8], since Patra et al. [20] provides the numbers for 10 bins only. The number of CE+TS [7] in parentheses is an optimal temperature tuned on a held-out set of Tiny-ImageNet.

| # of bins | 10 bins | | | 15 bins | | |
|---|---|---|---|---|---|---|
| Method | ACC↑ | ECE↓ | AECE↓ | ACC↑ | ECE↓ | AECE↓ |
| CE | 65.02 | 3.74 | 3.68 | 65.02 | 3.73 | 3.69 |
| CE+TS (1.1) [7] | 65.02 | 1.55 | 1.43 | 65.02 | 1.63 | 1.52 |
| MMCE [11] | 64.75 | 5.18 | 5.12 | 64.75 | 5.15 | 5.12 |
| CutMix [10] | 65.11 | 2.60 | 2.60 | 65.11 | 2.71 | 2.66 |
| CaPE-bin [15] | 59.78 | 2.91 | 2.61 | 59.78 | 2.99 | 2.96 |
| MCdrop [5] | **65.33** | 2.56 | 2.49 | **65.33** | 2.58 | 2.54 |
| Patra et al. [20] | 48.74 | 1.44 | - | 48.74 | - | - |
| ACLS (Ours) | 64.84 | **0.91** | **0.92** | 64.84 | **1.05** | **1.03** |



\end{document}